\documentclass[a4paper]{svproc}
\usepackage{makeidx}  
\makeindex

\usepackage{amsfonts}
\usepackage{multicol}
\usepackage{algpseudocode}
\usepackage{graphicx}
\usepackage{amsmath}
\usepackage[font=footnotesize]{caption}
\usepackage{cite}
\usepackage{subcaption}
\usepackage[table]{xcolor}
\usepackage[utf8]{inputenc}
\usepackage[bookmarks=false]{hyperref}

\definecolor{lightgray}{gray}{0.9}
\usepackage{url}

\makeatletter
\def\thanks#1{\protected@xdef\@thanks{\@thanks
        \protect\footnotetext{#1}}}
\makeatother

\author{Diogo Almeida\inst{1} \and Yiannis Karayiannidis\inst{2}
\thanks{This work is partially supported by the Swedish Foundation for
Strategic Research project GMT14-0082 FACT.}}
\institute{Division of Robotics, Perception and Learning, KTH Royal Institute of Technology, SE-100 44 Stockholm, Sweden,
\email{diogoa@kth.se}
\and
Dept. of Electrical Eng., Chalmers University of Technology, SE-412 96 Gothenburg, Sweden,
\email{yiannis@chalmers.se}}
\title{Asymmetric Dual-Arm Task Execution using an Extended Relative Jacobian}
\begin{document}
\maketitle
\begin{abstract}
	Coordinated dual-arm manipulation tasks can be broadly characterized as possessing absolute and relative motion components. 
	Relative motion tasks, in particular, are inherently redundant in the way they can be distributed between end-effectors. 
	In this work, we analyse cooperative manipulation in terms of the asymmetric resolution of relative motion tasks. 
	We discuss how existing approaches enable the asymmetric execution of a relative motion task, and show how an asymmetric relative motion space can be defined.
	We leverage this result to propose an extended relative Jacobian to model the cooperative system, which allows a user to set a concrete degree of asymmetry in the task execution. 
	This is achieved without the need for prescribing an absolute motion target. 
	Instead, the absolute motion remains available as a functional redundancy to the system.
	We illustrate the properties of our proposed Jacobian through numerical simulations of a novel differential Inverse~Kinematics algorithm.
\end{abstract}
\section{Introduction}
\setcounter{footnote}{0}
Consider a dual-armed robotic system, tasked with executing a \textit{coordinated} manipulation task. 
As opposed to \textit{uncoordinated} tasks, coordinated manipulation benefits from treating the two arms in the robotic system as a single kinematic chain. 
Concretely, the task space for coordinated manipulation can be defined in terms of \textit{absolute} and \textit{relative} motion components.
Absolute motion is equivalent to the resultant motion from external forces applied to a jointly carried object.
Relative motion, conversely, refers to the motion of one end-effector w.r.t the other \cite{Uchiyama1987, Uchiyama1988}.
This type of motion can be used to model tasks such as, e.g., assembly or tool manipulation.
The formulation of the manipulation task space in terms of absolute and relative motion is called the Cooperative Task Space (CTS) \cite{Chiacchio1996}.

We are interested in analysing the resolution of the CTS variables into end-effector velocities. 
Recent work proposes an Extended CTS (ECTS) definition which makes use of a novel definition of absolute motion to attain an \textit{asymmetric} resolution of the relative motion variable \cite{Park2015, Park2016}.
Alternatively, it is common to model solely the relative motion as a primary task quantity, by using the relative Jacobian formulation \cite{Lewis1990} when solving the dual-armed system's differential Inverse Kinematics (IK). 
In many cases, a secondary task is added which constrains the absolute motion of the system resulting in asymmetric relative motion \cite{Ajoudani2014,Jamisola2015, Hu2015, Foresi2017}.

In the following text, we discuss the CTS formulations and observe how a conflict between the definition of absolute and relative motion tasks is introduced by the ECTS method, Sec. \ref{sec_coop}.
In fact, ECTS relies in an asymmetric absolute motion definition, which we argue is at odds with the concept of absolute motion: commanding an 'asymmetric absolute motion' will result in the appearance of relative motion components. 
If we consider an absolute motion task such as jointly carrying an object, this will inevitably lead to internal forces on the object.
On the other hand, relative motion tasks are \textit{inherently redundant}.
The resolution of the relative motion task into the robot end-effectors can be arbitrary without any fundamental conflict with the concept of relative motion \cite{Almeida2018}.

We leverage our observations to propose an asymmetric relative motion space, which enables the asymmetric resolution of the relative motion without resorting to a redefinition of the absolute motion space, Sec. \ref{asymmetric_def}. 
The new space leads to the proposal of a novel relative Jacobian formulation and a corresponding differential IK algorithm, Sec. \ref{ik_sec}.
This allows a user to specify a concrete degree of asymmetry in the relative task resolution, within a relative Jacobian framework.
We illustrate the key properties of our approach through numerical simulations, Sec. \ref{case_studies}, and make our code freely available, which, in addition to the simulations, implements velocity controllers that employ all of the discussed methods\footnote{\url{https://github.com/diogoalmeida/asymmetric_manipulation}}.

\section{Related Work}
A common solution to the problem of bimanual cooperative manipulation is to employ a completely asymmetrical leader-follower (or master-slave) approach \cite{Lu1987, Zheng1989, Yan2016, Almeida2016, Almeida2016b}.
Alternatively, by considering the external and internal forces on a jointly held rigid object, the task can be modelled in terms of absolute and relative motion components \cite{Uchiyama1987, Uchiyama1988}.
The CTS definition results from a modelling approach which is independent of the statics of the dual-armed system \cite{Chiacchio1992, Chiacchio1996}.
The ECTS \cite{Park2015, Park2016} is obtained by redefining the absolute motion space of the coordinated system. 
CTS-based approaches have been used to describe coordinated tasks in, e.g., human-robot interaction settings \cite{Nemec2016}, the cooperative manipulation of a mechanism \cite{Almeida2018} or the execution of a bimanual dexterous manipulation task \cite{Cruciani2018}.

Relative Jacobian methods model solely the relative motion task \cite{Lewis1990, Lewis1996, Jamisola2015}, making it a common choice when addressing tasks that require just the relative motion space, such as machining \cite{Owen2005, Owen2008, Lee2012}, assembly \cite{Ajoudani2014} or drawing \cite{Lee2015}.
The absolute motion is part of the relative Jacobian's redundant space. 
This can be exploited, e.g., to enhance self-collision and obstacle avoidance as secondary tasks \cite{Mohri1996}, or for joint-limit avoidance \cite{Hu2015, Foresi2017, Ortenzi2018}. 

Hierarchical quadratic programs (HQP) \cite{Escande2014} are an alternative approach to obtain solutions to the problem of inverse differential kinematics.
While in this article we focus on pseudo-inverse solutions to the differential IK problem,  all the discussed Jacobian formulations can be employed in  the context of HQP, as showed in, e.g., \cite{Ceriani2016} and \cite{Tarbouriech2018} for the relative Jacobian.

\section{Cooperative motion spaces \label{sec_coop}}
In this section, we perform a task-space analysis of existing cooperation strategies. 
From the kinematic point of view, we will consider master-slave methods as examples of purely asymmetric task execution, which, as we will show, are covered by the analysed strategies.

CTS methods employ the full dual-arm task space, by defining an \textit{absolute} and a \textit{relative} motion spaces.
A possible approach to obtain \textit{asymmetric} relative motion is to redefine the absolute motion space, as is the case with ECTS.
We observe that this approach introduces conflicts between the relative and absolute motion tasks. 

\subsection{Notation}
Consider a dual-armed system composed by two robotic manipulators.
Let $\{h_i\}$ denote a generic coordinate frame, e.g., of the $i$-th manipulator's end-effector, where $i \in \{1, 2\}$. 
Each frame is defined by a position, $\mathbf{p}_i \in \mathbb{R}^3$ and orientation $\mathbf{R}_i \in \mathcal{SO}(3)$, expressed in a common base frame.
We write the angle-axis representation of $\mathbf{R}_i$ as $\mathbf{R}_{\mathbf{k}}(\vartheta_i)$, where $\vartheta_i$ is the angle one must rotate about the axis $\mathbf{k}$ to obtain $\mathbf{R}_i$.
The twist at each end-effector is defined as $\mathbf{v}_i = [\dot{\mathbf{p}}_i^\top\;\boldsymbol{\omega}_i^\top]^\top$, where $\boldsymbol{\omega}_i \in \mathbb{R}^3$ denotes the end-effectors' angular velocity.
Finally, we denote the nullspace of a matrix $\mathbf{A} \in \mathbb{R}^{n \times m}$ as $\mathcal{N}(\mathbf{A})$.

\subsection{Absolute and relative motion spaces}
The absolute and relative motion components can be derived from properly defined motion frames. 
\begin{figure}[t]
	\centering
	\begin{subfigure}[b]{0.35\textwidth}
		\includegraphics[width = \textwidth]{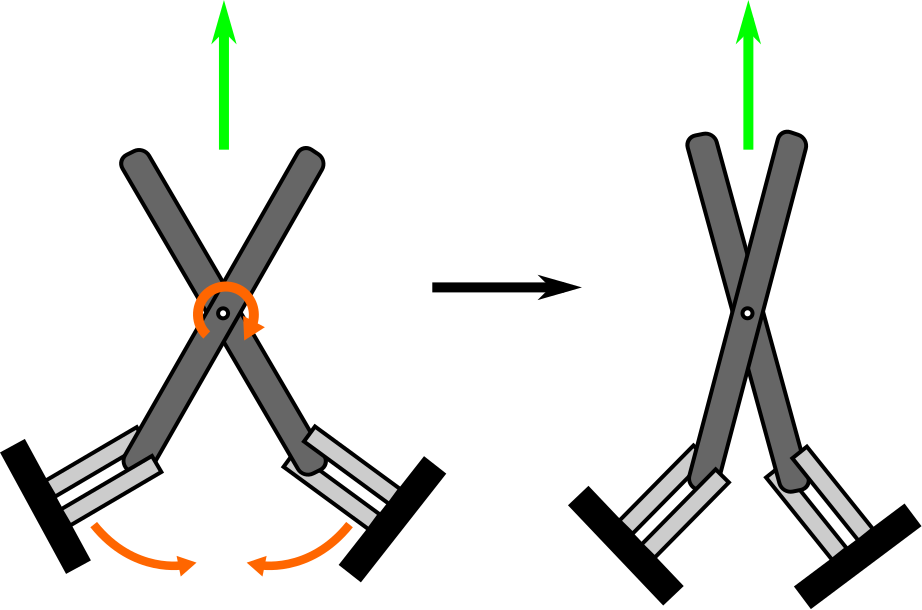}
		\caption{\textbf{Symmetric} execution.}
	\end{subfigure}
	\quad
	\begin{subfigure}[b]{0.35\textwidth}
		\includegraphics[width = \textwidth]{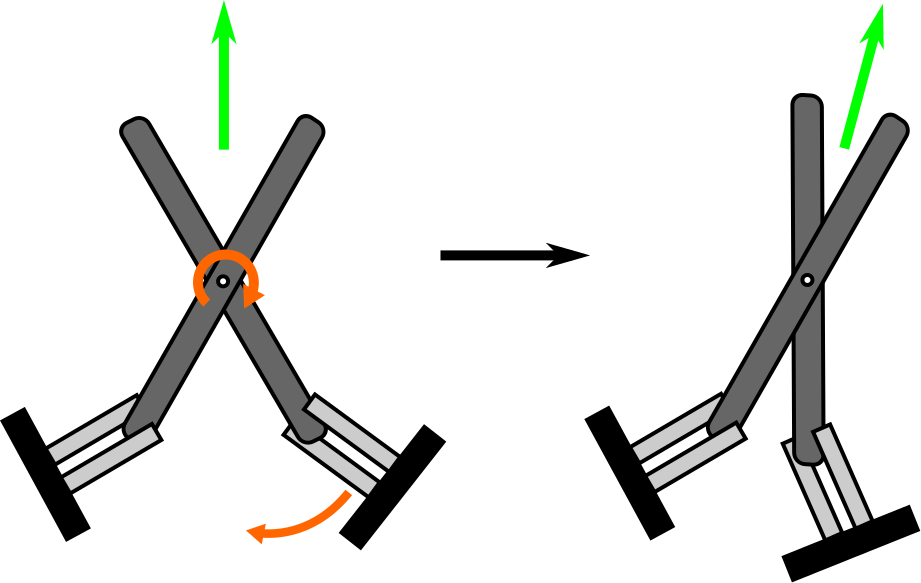}
		\caption{\textbf{Asymmetric} execution}
	\end{subfigure}
	\caption{A simple relative motion task is to open/close a pair of scissors. Note that an asymmetric execution results in a change of the average orientation (green arrow) of the two scissors' pieces. \label{asymmetric_motion_fig}}
\end{figure}
\subsubsection{Cooperative Task Space \label{cts}}
In CTS \cite{Chiacchio1996}, absolute and relative motion frames are defined, respectively $\{h_a\}$ and $\{h_r\}$, such that
\begin{equation}
  \mathbf{p}_a = \frac{1}{2}(\mathbf{p}_1 + \mathbf{p}_2) \quad\text{and}\quad \mathbf{R}_a = \mathbf{R}_1\mathbf{R}_{\mathbf{k}_{1,2}}\left (\frac{\vartheta_{1,2}}{2}\right),
  \label{cts_abs}
\end{equation}  
where $\mathbf{k}_{1,2}$ and $\vartheta_{1,2}$ are extracted from the angle-axis representation of ${^1}\mathbf{R}_2 = \mathbf{R}_1^\top \mathbf{R}_2$, and
\begin{equation}
  \mathbf{p}_r = \mathbf{p}_2 - \mathbf{p}_1, \quad \mathbf{R}_r = {^1}\mathbf{R}_2.
  \label{cts_rel}
\end{equation}
The absolute and relative motion twists, respectively $\mathbf{v}_a$ and $\mathbf{v}_r$ can be obtained through differentiation,
\begin{equation}
	\mathbf{v}_a = \frac{1}{2}(\mathbf{v}_1 + \mathbf{v}_2) \quad\text{and}\quad \mathbf{v}_r = \mathbf{v}_2 - \mathbf{v}_1.
\end{equation}
In task space, this relation can be expressed by a \textit{linking} matrix, $\mathbf{L}_{cts} \in \mathbb{R}^{12}$.
Let $\mathbf{v}_{cts} = [\mathbf{v}_a^\top\; \mathbf{v}_r^\top]^\top$ and $\mathbf{v} = [\mathbf{v}_1^\top\;\mathbf{v}_2^\top]^\top$. 
Then,
\begin{equation}
	 \mathbf{v}_{cts} = \mathbf{L}_{cts} \mathbf{v}, \quad \mathbf{L}_{cts} = \begin{bmatrix}
		\frac{1}{2}\mathbf{I}_6 & \frac{1}{2}\mathbf{I}_6\\
		-\mathbf{I}_6 & \mathbf{I}_6
	\end{bmatrix}.
	\label{cts_link}
\end{equation}
The CTS linking matrix $\mathbf{L}_{cts}$ is square and nonsingular, and $\mathbf{v}$ can be recovered through matrix inversion,
\begin{equation}
	\begin{bmatrix}
		\mathbf{v}_1\\
		\mathbf{v}_2
	\end{bmatrix} = \begin{bmatrix}
		\mathbf{I}_6 & -\frac{1}{2}\mathbf{I}_6\\
		\mathbf{I}_6 & \frac{1}{2} \mathbf{I}_6
	\end{bmatrix} \begin{bmatrix}
		\mathbf{v}_a\\
		\mathbf{v}_r
	\end{bmatrix}.
	\label{cts_solution}
\end{equation}
It is clear from \eqref{cts_solution} that the coordinated task is divided \textit{symmetrically} between the two end-effectors: each end-effector executes the prescribed absolute motion, and the relative motion task is divided evenly among them.

\subsubsection{Extended Cooperative Task Space \label{ects}}
The CTS formulation has been extended in \cite{Park2015, Park2016}. A different definition for the absolute frame is adopted,
\begin{equation}
	\mathbf{p}_a = \alpha \mathbf{p}_1 + (1 - \alpha) \mathbf{p}_2, \qquad \mathbf{R}_a =  \mathbf{R}_1\mathbf{R}_{\mathbf{k}_{1,2}}\left((1 - \alpha)\vartheta_{1,2}\right),
	\label{ects_abs}
\end{equation}
with  $0 \leq \alpha \leq 1$.
The ECTS linking matrix is given by
\begin{equation}
	\mathbf{L}_{E}(\alpha) = \begin{bmatrix}
		\alpha \mathbf{I}_6 & (1 - \alpha) \mathbf{I}_6\\
		- \mathbf{I}_6 & \mathbf{I}_6,
	\end{bmatrix}
	\label{ects_linking}
\end{equation}
and $\mathbf{v}_{cts} = \mathbf{L}_{E}(\alpha)\mathbf{v}$, as in \eqref{cts_link}.
The effect of the cooperation parameter $\alpha$ is clear through the inversion of \eqref{ects_linking},
\begin{equation}
	\mathbf{L}_E(\alpha)^{-1} = \begin{bmatrix}
		\mathbf{I}_6 & - (1 - \alpha) \mathbf{I}_6\\
		\mathbf{I}_6 & \alpha \mathbf{I}_6
	\end{bmatrix},
	\label{inverse_ects_linking}
\end{equation}
That is, by setting $\alpha$, asymmetric relative motion can be achieved, such that $\mathbf{v}_a = \mathbf{0}$ in the new absolute frame.
The relative motion task can be solved in a serial (master-slave, $\alpha = 0$ or $\alpha = 1$), blended (asymmetrical, $\alpha \neq 0.5$) or parallel (symmetrical, $\alpha = 0.5$) mode of cooperation.
Note that the introduction of asymmetries in the relative motion affects the symmetric absolute pose, Fig.~\ref{asymmetric_motion_fig}.

Both $\mathbf{L}_{cts}$ and $\mathbf{L}_E(\alpha)$ are composed by an absolute and a relative part,
\begin{equation}
	\mathbf{L}_{cts} = \begin{bmatrix}
		\mathbf{L}_a\\
		\mathbf{L}_r
	\end{bmatrix} \quad \mathbf{L}_{E}(\alpha) = \begin{bmatrix}
		\boldsymbol{\mathcal{L}}_{a}(\alpha)\\
		\mathbf{L}_r
	\end{bmatrix},
	\label{linking_matrices}
\end{equation}
where $\mathbf{L}_a, \mathbf{L}_r, \boldsymbol{\mathcal{L}}_a(\alpha) \in \mathbb{R}^{6 \times 12}$.
These matrices depend on the definition of the frames where the absolute and relative motion are expressed.
For both CTS and ECTS, the relative motion is given by $\mathbf{v}_r = \mathbf{L}_r \mathbf{v}$.
CTS adopts the symmetric resolution of the absolute motion, $\mathbf{v}_a = \mathbf{L}_a \mathbf{v}$ while ECTS makes use of an asymmetric formulation, $\mathbf{v}_a =\boldsymbol{\mathcal{L}}_a(\alpha)\mathbf{v}$. 

\begin{figure*}[t]
	\centering
	\begin{subfigure}[b]{0.25\textwidth}
		\centering
		\includegraphics[width = \textwidth]{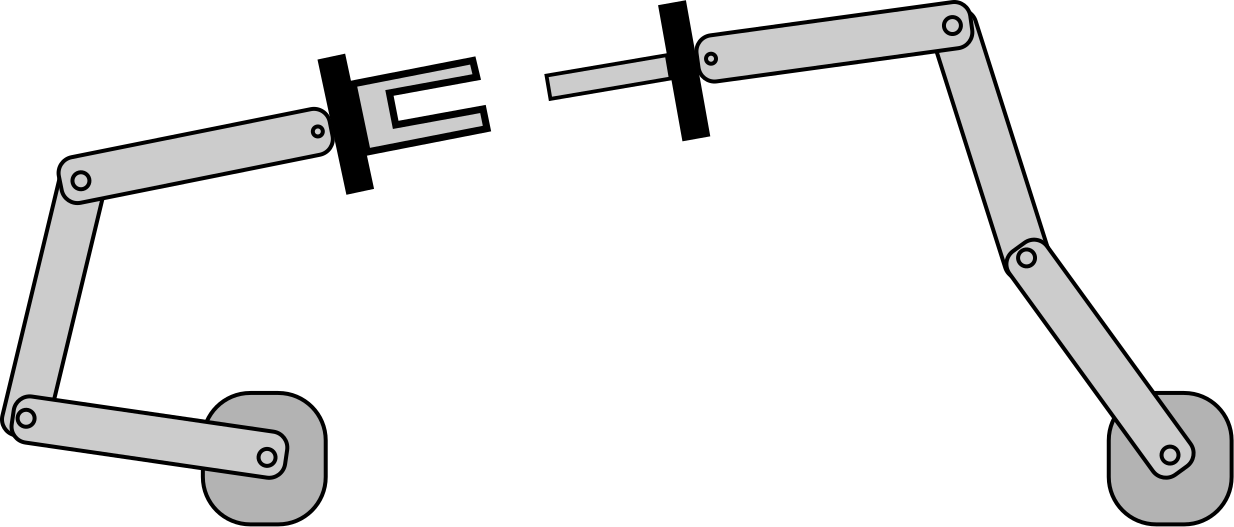}
		\caption{}
	\end{subfigure}
	\quad
	\begin{subfigure}[b]{0.25\textwidth}
		\centering
		\includegraphics[width = \textwidth]{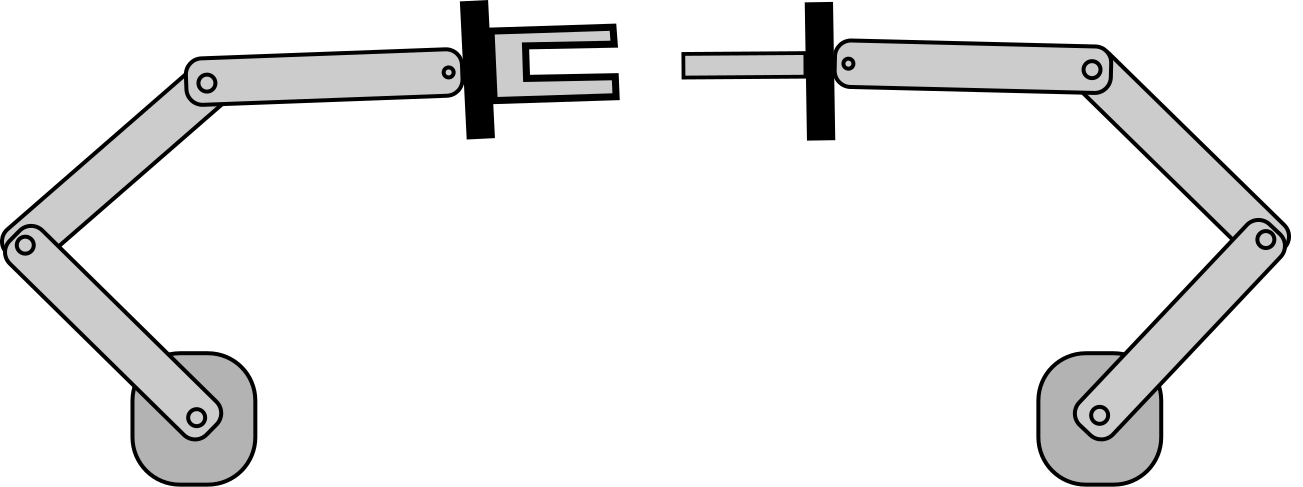}
		\caption{}
	\end{subfigure}
	\quad
	\begin{subfigure}[b]{0.25\textwidth}
		\centering
		\includegraphics[width = \textwidth]{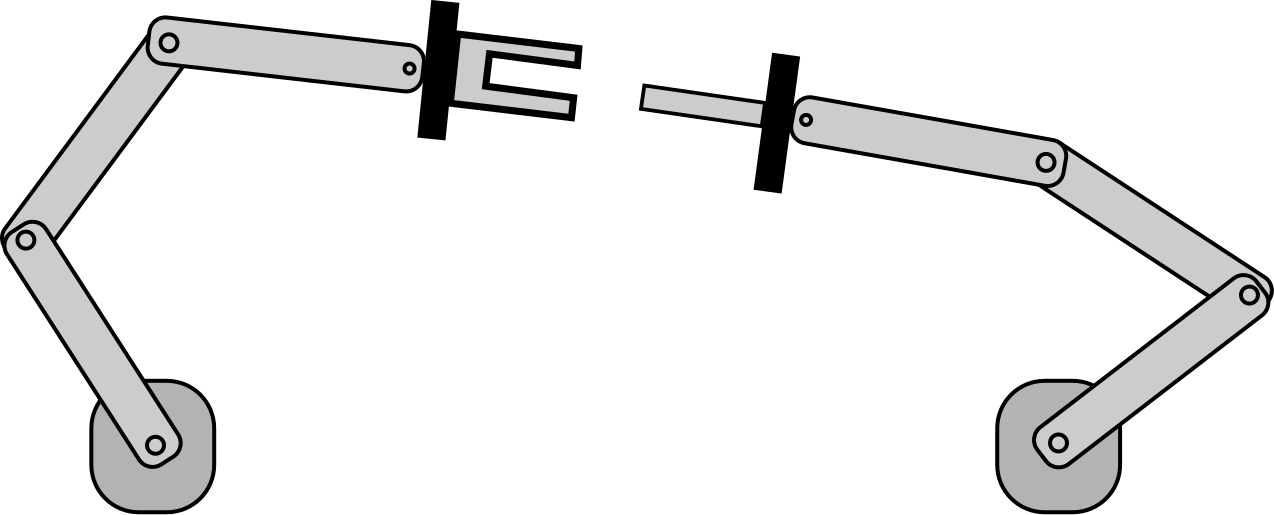}
		\caption{}
	\end{subfigure}
	\caption{The absolute motion of the cooperative system is redundant w.r.t the relative motion task.\label{absolute_redundancy} }
\end{figure*}

The (E)CTS motion space fully specifies the cooperative motion in terms of absolute and relative variables. 
In many tasks, however, this overconstrains the problem.
For example, a peg-in-hole assembly can be executed by having each robot arm grasp a part and executing a relative motion between its end-effectors. 
The absolute motion is a functional redundancy in this case, Fig. \ref{absolute_redundancy}.
Other examples include machining \cite{Owen2005} or drawing \cite{Lee2015}.

\subsubsection{Relative motion space \label{rel_space}}
For tasks which can be solved exclusively through relative motion, it is possible to specify a desired $\mathbf{v}_r$ only.
This removes the need to adopt the $12 \times 12$ dimensional linking matrices in \eqref{linking_matrices}. 
In this scenario, we can use the Moore-Penrose pseudo-inverse, $\mathbf{L}_r^\dagger = \mathbf{L}_r^\top (\mathbf{L}_r\mathbf{L}_r^\top)^{-1}$, to resolve a desired relative motion into velocities of each robot end-effector, 
\begin{equation}
	\mathbf{v} = \mathbf{L}_r^\dagger \mathbf{v}_r = \frac{1}{2}\begin{bmatrix}
		-\mathbf{I}_6\\
		\mathbf{I}_6
	\end{bmatrix}\mathbf{v}_r,
	\label{symmetric_relative_solution}
\end{equation}
which matches the relative part of the CTS solution \eqref{cts_solution}.
An homogenous solution can be added through a nullspace projection, e.g.,
\begin{equation}
			\begin{aligned}
				\mathbf{v} &= \mathbf{L}_r^\dagger\mathbf{v}_r + (\mathbf{I}_{12} - \mathbf{L}_r^\dagger\mathbf{L}_r)[\mathbf{I}_6\quad \mathbf{0}_6]^	\dagger\mathbf{v}_{1_d}\\
				& = \frac{1}{2} \begin{bmatrix}
					 \mathbf{v}_{1_d} - \mathbf{v}_r\\
					 \mathbf{v}_{1_d} + \mathbf{v}_r
				\end{bmatrix} = \frac{1}{2}\begin{bmatrix}
					 \mathbf{I}_6 & -\mathbf{I}_6\\
					\mathbf{I}_6 & \mathbf{I}_6
				\end{bmatrix} \begin{bmatrix}
					\mathbf{v}_{1_d}\\
					\mathbf{v}_r
				\end{bmatrix}
			\end{aligned}.
			\label{v1_sec}
\end{equation}
In this solution, a secondary task $\{h_1\}$ is being mapped to the absolute motion space, with $\mathbf{v}_a = \frac{1}{2} \mathbf{v}_{1_d}$. 
The resulting velocity distribution is analogous to the CTS case, eq. \eqref{cts_solution}.
However, if the secondary objective is not prescribed in $\{h_a\}$, a conflict between primary and secondary tasks is introduced which will induce asymmetries in the motion of the system's end-effectors.

\begin{figure}[t]
	\centering
	\includegraphics[width = 0.9\textwidth]{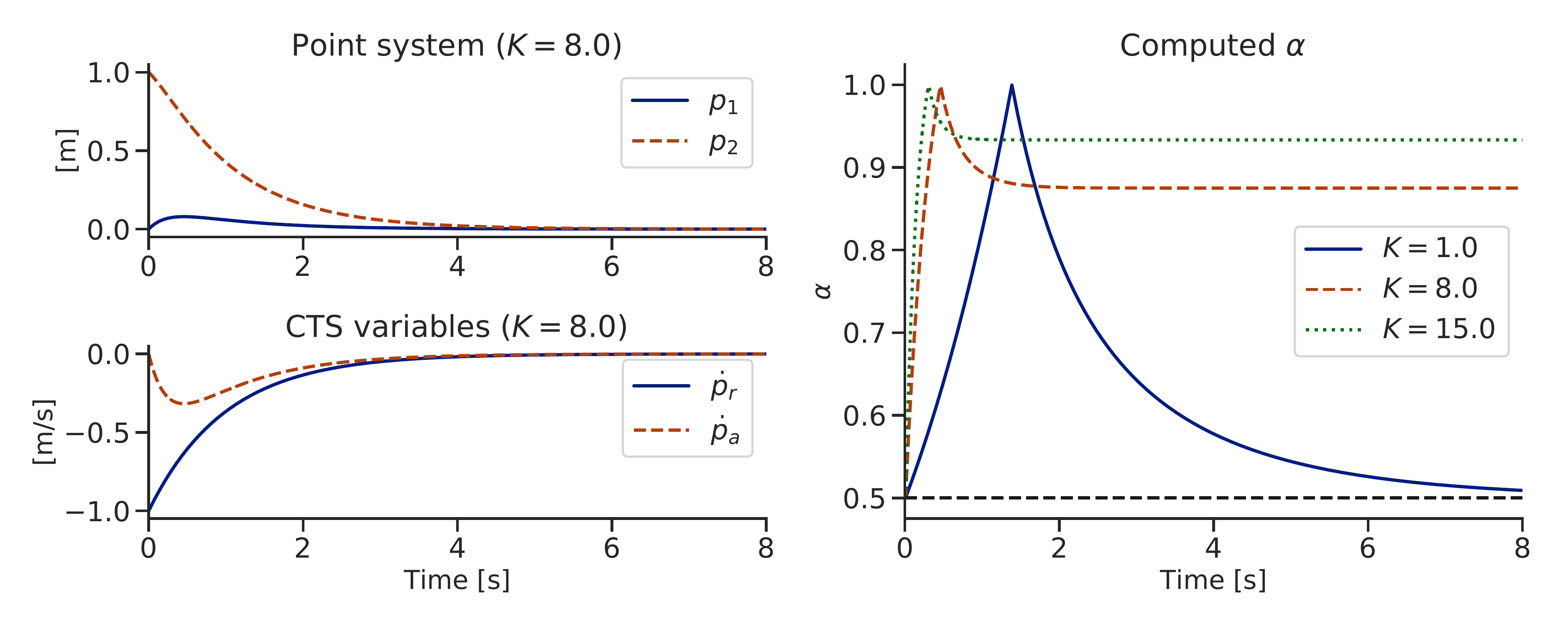}
	\caption{Numerical simulation of the point-system \eqref{point_system}. Different values of $K$ affect the degree of asymmetry $\alpha$ in the execution of the relative motion task.\label{1_d_asym_reljac}}
\end{figure}

\begin{example}[Constructing asymmetric relative motion]
	Let $p_1, p_2 \in \mathbb{R}$, with initial state $p_1 = 0$ and $p_2 = 1$, and consider the relative velocity command $\dot{p}_{r_d} = p_1 - p_2$. 
	Additionally, let $\dot{p}_{1_d} = -Kp_1$, with $K > 0$.
	The solution \eqref{v1_sec}, applied to this one-dimensional case, yields the autonomous system
	\begin{equation}
		\begin{bmatrix}
			\dot{p}_1\\
			\dot{p}_2
		\end{bmatrix} = \frac{1}{2}\begin{bmatrix}
			-(K + 1) & 1\\
			1 - K & -1
		\end{bmatrix}\begin{bmatrix}
			p_1\\
			p_2
		\end{bmatrix}.
		\label{point_system}
	\end{equation}		
	The relative velocity of this system is given by $\dot{p}_r = \dot{p}_2 - \dot{p}_1 = \dot{p}_{r_d}$, i.e., the desired relative command is attained.
	The degree of asymmetry in the motion of the two points can be computed through $\alpha = \frac{|\dot{p}_2|}{|\dot{p}_1| + |\dot{p}_2|}$ and will depend on the secondary task gain $K$. 
	We depict the time evolution of system \eqref{point_system}, as well as $\alpha$ for different values of $K$ in Fig. \ref{1_d_asym_reljac}. 
	It is clear that for a larger $K$ the more asymmetric is the motion of the two points, since the induced absolute motion will regulate $p_1$ to remain at the origin. 
	For $K = 8$ in particular, it can be seen that the injected absolute motion $\dot{p}_a$ allows $p_2$ to contribute the most to the relative motion task.
	\label{example_mapping}
\end{example}

\subsection{Asymmetric resolution of the CTS variables \label{asym_motion}}
We have seen in Sec. \ref{cts} that inverting the mapping \eqref{cts_link} yields a symmetric distribution of the CTS variables into the two robot end-effectors.
In the current literature, two predominant coordination approaches achieve an asymmetric execution of the CTS variables:
\begin{enumerate}
	\item In the ECTS \cite{Park2015, Park2016}, a redefined absolute motion space constrains the resolution of the relative motion to be asymmetric, Sec. \ref{ects}
	\item Methods which consider only the relative motion space often add a secondary task to one of the end-effectors \cite{ Ajoudani2014,Jamisola2015, Hu2015, Foresi2017}, which results in asymmetric behavior, see Example \ref{example_mapping}
\end{enumerate}
Both methods combine the CTS variables to achieve asymmetric behavior. 
The ECTS solution for the absolute motion introduces relative motion components,
\begin{equation}
	\mathbf{v}_r = \mathbf{L}_r \boldsymbol{\mathcal{L}}_{a}(\alpha)^\dagger \mathbf{v}_a = \frac{1 - 2\alpha}{\alpha^2 + (1 - \alpha)^2}\mathbf{v}_a,
	\label{relative_from_abs}
\end{equation}
while the nullspace projection \eqref{v1_sec} maps the secondary task of the system to the absolute motion space.
We will show how this projection can be modified to introduce a deliberate degree of asymmetry in the execution of the relative motion task.

\section{Defining an asymmetric relative motion space \label{asymmetric_def} }
Consider that we wish to control the relative motion between the end-effectors in an \textit{asymmetric} manner. 
Analogously to the ECTS approach, we would like to be able to specify a degree of cooperation between the arms.
This can be achieved through a redefined relative motion frame.
If we denote the angle-axis representations of $\mathbf{R}_1 = \mathbf{R}_{\mathbf{k}_1}(\vartheta_1)$ and $\mathbf{R}_2 = \mathbf{R}_{\mathbf{k}_2}(\vartheta_2)$, then the asymmetric relative motion frame $\{h_r\}$ can be defined as
\begin{equation}
	\begin{aligned}
		\mathbf{p}_r &= \frac{\alpha\mathbf{p}_2 - (1 - \alpha)\mathbf{p}_1}{(1 - \alpha)^2 + \alpha^2}\\
		\mathbf{R}_r &= \mathbf{R}_{\mathbf{k}_1}^\top\left(\frac{(1 - \alpha)\vartheta_1}{(1 - \alpha)^2 + \alpha^2}\right)\mathbf{R}_{\mathbf{k}_2}\left(\frac{\alpha\vartheta_2}{(1 - \alpha)^2 + \alpha^2}\right).
	\end{aligned}
	\label{asymmetric_relative_frame}
\end{equation}
Asymmetric relative motion can be obtained through the differentiation of \eqref{asymmetric_relative_frame}, 
\begin{equation}
	\mathbf{v}_r = \boldsymbol{\mathcal{L}}_r(\alpha) \mathbf{v},
	\label{asym_relative_link}
\end{equation}
where the asymmetric relative linking matrix $\boldsymbol{\mathcal{L}}_r(\alpha)$ is
\begin{equation}
	\boldsymbol{\mathcal{L}}_r (\alpha) = \frac{1}{(1 - \alpha)^2 + \alpha^2} \begin{bmatrix}
		-(1 - \alpha)\mathbf{I}_6 & \alpha \mathbf{I}_6
	\end{bmatrix}.
	\label{asymmetric_linking}
\end{equation}
The particular solution to \eqref{asym_relative_link} is analogous to \eqref{symmetric_relative_solution} and corresponds to the the ECTS asymmetric distribution of the relative motion\eqref{inverse_ects_linking} ,
\begin{equation}
	\mathbf{v} = \boldsymbol{\mathcal{L}}_r(\alpha)^\dagger \mathbf{v}_r = \begin{bmatrix}
		-(1 - \alpha) \mathbf{I}_6\\
		\alpha \mathbf{I}_6
	\end{bmatrix}\mathbf{v}_r.
	\label{asymmetric_inv}
\end{equation}

\begin{theorem}
	The motion space defined by $\{h_r\}$ in \eqref{asymmetric_relative_frame} is characterized by the following properties:
	\begin{enumerate}
		\item Commanding a relative velocity in $\{h_r\}$, eq. \eqref{asymmetric_inv}, renders the asymmetric absolute motion space \eqref{ects_abs} invariant.
		\item The inverse mapping $\boldsymbol{\mathcal{L}}_r(\alpha)^\dagger$ is a generalized inverse of $\mathbf{L}_r$.
	\end{enumerate}
	\begin{proof}
		Let
		\begin{equation}
			\mathbf{v}_a = \boldsymbol{\mathcal{L}}_a(\alpha) \mathbf{v},
		\end{equation}
		and consider the solution \eqref{asymmetric_inv}.
		We have,
		\begin{equation}
			\mathbf{v}_a =  \boldsymbol{\mathcal{L}}_a(\alpha)\boldsymbol{\mathcal{L}}_r(\alpha)^\dagger\mathbf{v}_r = \mathbf{0},
			\label{asymmetric_orthogonality}
		\end{equation}
		and thus, the asymmetric absolute motion frame \eqref{ects_abs} remains invariant to velocities commanded in the motion frame from \eqref{asymmetric_relative_frame}.
		$\boldsymbol{\mathcal{L}}_r(\alpha)$ acts as a generalized inverse to $\mathbf{L}_r$ since
		\begin{equation}
			\mathbf{L}_r\boldsymbol{\mathcal{L}}_r(\alpha)^\dagger = \begin{bmatrix}
				-\mathbf{I}_6 & \mathbf{I}_6
			\end{bmatrix} \begin{bmatrix}
				-(1 - \alpha) \mathbf{I}_6\\
				\alpha \mathbf{I}_6
			\end{bmatrix} = \mathbf{I}_6.
			\label{generalized_inverse}
		\end{equation}
	\end{proof}
	\label{asym_theo}
\end{theorem}


\begin{corollary}
	The solution for the asymmetric relative motion space \eqref{asymmetric_inv} can be obtained by adding an homogeneous component to \eqref{symmetric_relative_solution}. 
	\begin{proof}
		This follows straightforwardely from the fact that $\boldsymbol{\mathcal{L}}_r(\alpha)$ acts as a generalized inverse to $\mathbf{L}_r$\eqref{generalized_inverse},
		\begin{equation}
			\mathbf{v} = \mathbf{L}_r^\dagger \mathbf{v}_r + (\mathbf{I}_{12} - \mathbf{L}_r^\dagger \mathbf{L}_r)\boldsymbol{\mathcal{L}}	_r(\alpha)^\dagger \mathbf{v}_r = \boldsymbol{\mathcal{L}}_r(\alpha)^\dagger\mathbf{v}_r.
			\label{task_priority_asym}
		\end{equation}
	\end{proof}
	\label{asym_corollary}
\end{corollary}
The solution \eqref{task_priority_asym} generates the required absolute motion to ensure the degree of asymmetry $\alpha$ in the execution of the prescribed relative velocity $\mathbf{v}_r$,
\begin{equation}
	\mathbf{v}_a = \mathbf{L}_a \boldsymbol{\mathcal{L}}_r(\alpha)^{\dagger} \mathbf{v}_r = \frac{2\alpha - 1}{2}\mathbf{v}_r.
	\label{induced_sym_abs}
\end{equation}
We can use this result to construct a relative Jacobian method for solving the system's differential IK which takes into account a desired degree of asymmetry.

\section{Differential Inverse Kinematics \label{ik_sec}}
The prescribed cooperative motion can be distributed between the two end-effectors, as seen in the previous section, and each arm can solve its differential IK separately.
Alternatively, we can derive differential IK algorithms which directly resolve a desired cooperative motion to the joint state of the complete dual-arm chain.

\subsection{Notation}
Let $\mathbf{q}_i \in \mathbb{R}^n$ represent the joint variables of the $i$-th manipulator, with $n \geq 6$, and $\mathbf{J}_i(\mathbf{q}_i) \in \mathbb{R}^{6\times n}$ its Jacobian, such that $\mathbf{v}_i = \mathbf{J}_i(\mathbf{q}_i)\dot{\mathbf{q}}_i$.
We will omit the Jacobian's dependency on the joint variables for the remaining of this text, and assume that the manipulators' Jacobians are full rank, i.e., the manipulators are not operating in a singular configuration.
Finally, let $\mathbf{S}(\mathbf{a}) \in \mathbb{R}^{3 \times 3}$ be the skew-symmetric matrix such that, for $\mathbf{a}, \mathbf{b} \in \mathbb{R}^3$, $\mathbf{S}(\mathbf{a})\mathbf{b} = \mathbf{a} \times \mathbf{b}$.

\subsection{Joint Jacobians}
Let $\mathbf{J} = \begin{bmatrix}
	\mathbf{J}_1 & \mathbf{0}_{6 \times n}\\
	\mathbf{0}_{6 \times n} & \mathbf{J}_2
\end{bmatrix}$.
We can solve the cooperative task by using any of the methods in section \ref{sec_coop} and computing the differential IK for each individual manipulator. 
As an example, a relative motion task can be solved into joint space by computing
\begin{equation}
	\dot{\mathbf{q}} = \mathbf{J}^\dagger \mathbf{L}_r^\dagger \mathbf{v}_r,
	\label{ik_individual}
\end{equation}
where $\mathbf{q} = [\mathbf{q}_1^\top,\;\mathbf{q}_2^\top]^\top$. 
The nullspace of $\mathbf{J}$ has dimension $\text{dim}(\mathcal{N}(\mathbf{J})) = 2n - 12$.
Alternatively, all the previously discussed cooperative motion definitions can be extended to a mapping from task to joint spaces through the construction of joint Jacobians. 
This is equivalent to treating the two manipulators as a single kinematic chain.
We can obtain the joint Jacobians through a pre-multiplication with the appropriate linking matrix, e.g., $\mathbf{J}_{cts} = \mathbf{L}_{cts}\mathbf{J} \in \mathbb{R}^{12\times 2n}$.
It is often convenient, however, to express the cooperative task w.r.t object frames rigidly connected to each end-effector, $\{h_{o_i}\}$. 
These can represent, e.g., the tip of the peg and the center of the hole in a peg-in-hole type of assembly task.
In this case, a transformation between the twists at $\{h_{o_i}\}$ and the corresponding $\{h_i\}$ is needed.

We define two virtual sticks, which connect $\mathbf{p}_{o_i}$ to the end-effector's positions $\mathbf{p}_i$, such that $\mathbf{r}_i = \mathbf{p}_{o_i} - \mathbf{p}_i$. 
The necessary screw transformation is defined as
$
	\mathbf{W}_i = \begin{bmatrix}
 		\mathbf{I}_3 & -\mathbf{S}(\mathbf{r}_i)\\
		\mathbf{O}_3 & \mathbf{I}_3
	\end{bmatrix},
$
such that $\mathbf{v}_{o_i} = \mathbf{W}_i \mathbf{v}_{i}$.
Let $\mathbf{W} = \begin{bmatrix}
	\mathbf{W}_1 & \mathbf{0}_6\\
	\mathbf{0}_6 & \mathbf{W}_2
\end{bmatrix}$. 
The joint Jacobians are
\begin{equation}
	\mathbf{J}_{cts} = \mathbf{L}_{cts}\mathbf{W}\mathbf{J} \qquad \mathbf{J}_E(\alpha) = \mathbf{L}_{E}(\alpha)\mathbf{W}\mathbf{J},
	\label{cts_jacobians}
\end{equation}
for the CTS formulations, where $\mathbf{J}_{cts},\, \mathbf{J}_E(\alpha) \in \mathbb{R}^{12 \times 2n}$ and
\begin{equation}
	\mathbf{J}_r = \mathbf{L}_r \mathbf{W}\mathbf{J},
	\label{relative_jacobian}
\end{equation}
in case only a relative motion task is specified, with $\mathbf{J}_r \in \mathbb{R}^{6 \times 2n}$. 
Note that, in general, $\mathbf{J}_r^\dagger \neq \mathbf{J}^\dagger \mathbf{W}^\dagger \mathbf{L}_r^\dagger$.
In fact, treating the dual-armed system as a single kinematic chain results, in general, in a larger dimension of the Jacobian nullspace, as $\text{rank}(\mathbf{J}_r) \leq 6$ and thus
\begin{equation}
	\text{dim}(\mathcal{N}(\mathbf{J}_r)) \geq \text{dim}(\mathcal{N}(\mathbf{J})) =  \text{dim}(\mathcal{N}(\mathbf{J}_1)) + \text{dim}(\mathcal{N}(\mathbf{J}_2)).
\end{equation} 
In practice, the larger nullspace includes absolute motion components which are not present when solving \eqref{ik_individual}.
The Jacobian in \eqref{relative_jacobian} is often called relative Jacobian.

The forward differential kinematics for a relative motion task are expressed as
\begin{equation}
	\mathbf{v}_r = \mathbf{J}_r \dot{\mathbf{q}}.
	\label{fk_relative}
\end{equation}
In the following, we will focus on finding feasible joint space solutions for \eqref{fk_relative}.
It is well known that such solutions have the general form
\begin{equation}
	\dot{\mathbf{q}} = \mathbf{J}_r^\dagger\mathbf{v}_r + (\mathbf{I}_{2n} - \mathbf{J}_r^\dagger\mathbf{J}_r)\boldsymbol{\zeta},
	\label{ik}
\end{equation}
with $(\mathbf{I}_{2n} - \mathbf{J}_r^\dagger\mathbf{J}_r)\boldsymbol{\zeta}$ being part of the homogeneous solution to \eqref{fk_relative}.
The joint space vector $\boldsymbol{\zeta} \in \mathbb{R}^{2n}$ composes a secondary task, added to $\mathbf{v}_r$ in a task-priority manner.
By setting $\boldsymbol{\zeta} = \mathbf{0}$ we get the minimum norm solution for $\dot{\mathbf{q}}$.
The symmetric absolute motion space is orthogonal to the relative motion space used in the relative Jacobian definition \eqref{relative_jacobian},
\begin{equation}
	\mathbf{L}_r \mathbf{L}_a^\dagger =  \mathbf{0}.
	\label{symmetric_orthogonality}
\end{equation}
This absolute motion is then a functional redundancy to \eqref{fk_relative} and thus, in general, its minimum norm solution can contain absolute motion components, depending on the manipulators' design.
The desirability of this property depends on the task requirements.
If the absolute motion must be prescribed, using the CTS is ideal.
In light of the discussion in Sec. \ref{asym_motion}, we argue that the ECTS can be used to achieve asymmetric relative motion when $\mathbf{v}_a = \mathbf{0}$. 
However, if $\mathbf{v}_a \neq 0$ is to be commanded, ECTS requires $\alpha = 0.5$\footnote{In fact, this is the case in \cite{Park2016}.}.
Otherwise, the resulting asymmetric absolute motion command will contain relative motion components, which is undesirable.
Alternatively, using \eqref{ik_individual} prevents any non-specified absolute motion from occurring. 
When it is acceptable for absolute motion to be exploited as a functional redundancy, exploitation strategies include, e.g., workspace obstacle and self-collisions avoidance \cite{Mohri1996}.

\subsection{Asymmetric relative Jacobian}
\def \robwidth{0.2\textwidth}
\begin{figure}[t]
	\centering
	\begin{subfigure}[b]{\robwidth}
		\includegraphics[width = \textwidth, trim={300, 200, 200, 100}, clip]{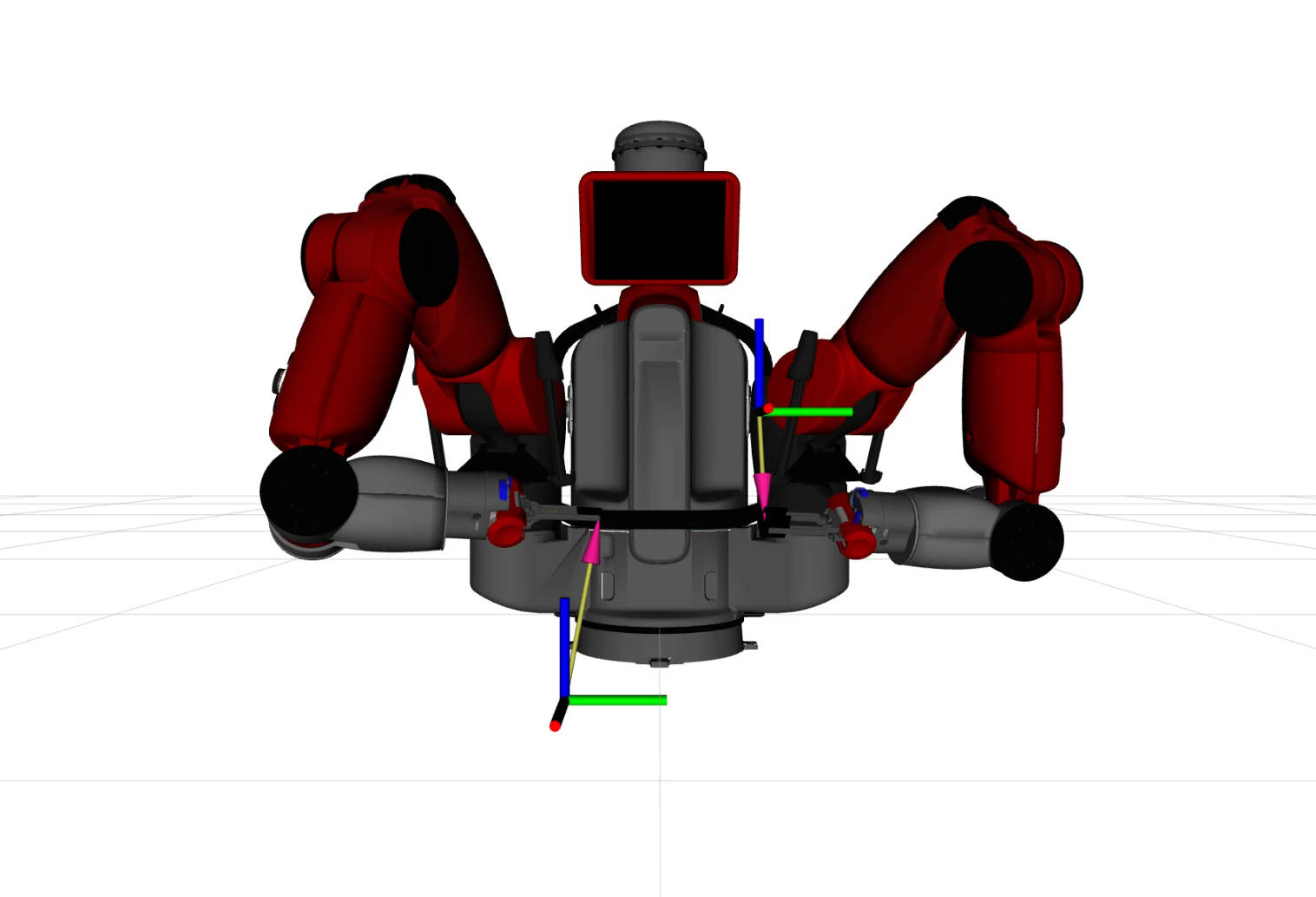}
		\caption{\label{linear_front}}
	\end{subfigure}
	\quad
	\begin{subfigure}[b]{\robwidth}
		\includegraphics[width = \textwidth, trim={200, 200, 100, 100}, clip]{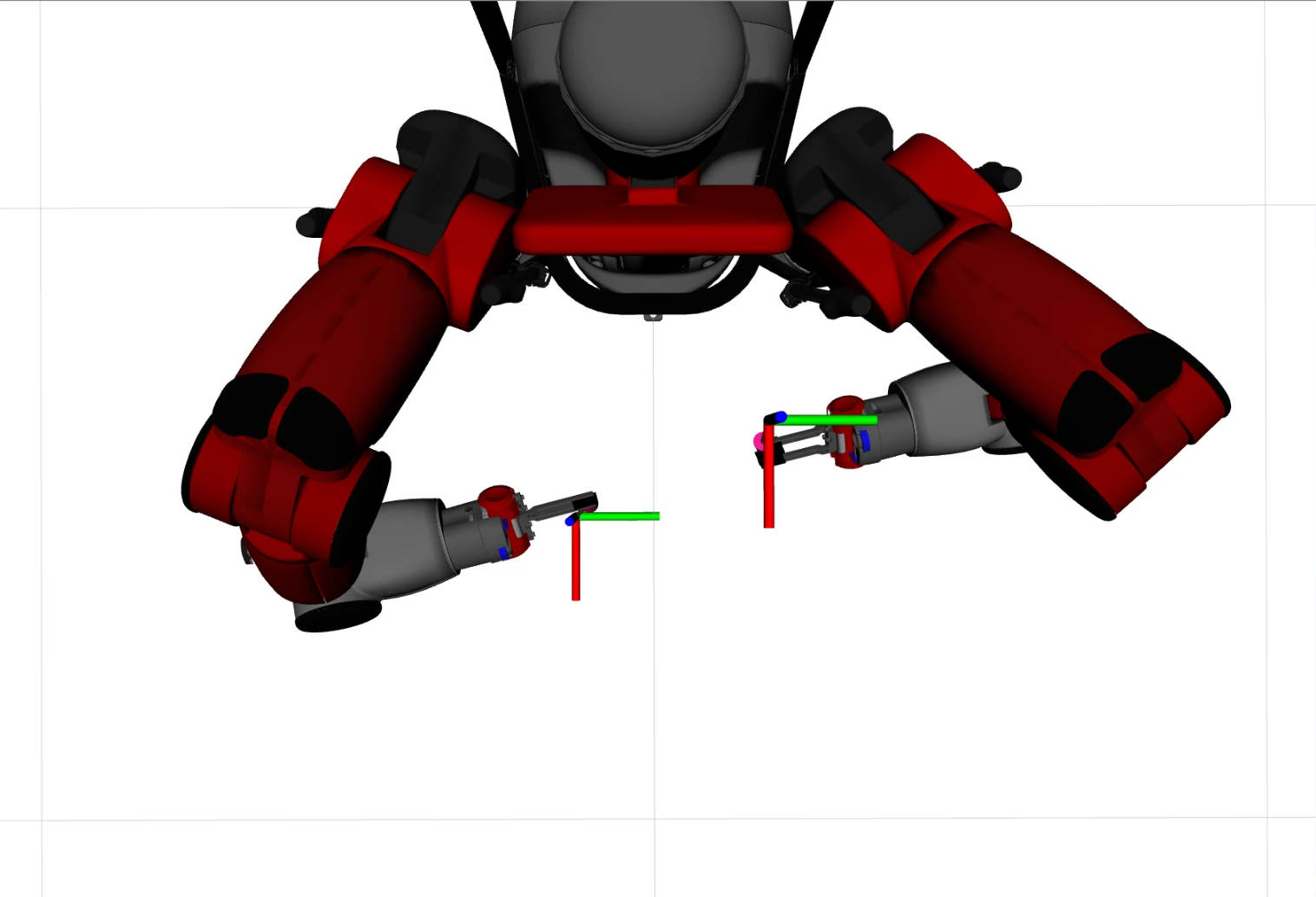}
		\caption{\label{linear_top}}
	\end{subfigure}
	\qquad
	\begin{subfigure}[b]{\robwidth}
		\includegraphics[width = \textwidth, trim={300, 200, 200, 100}, clip]{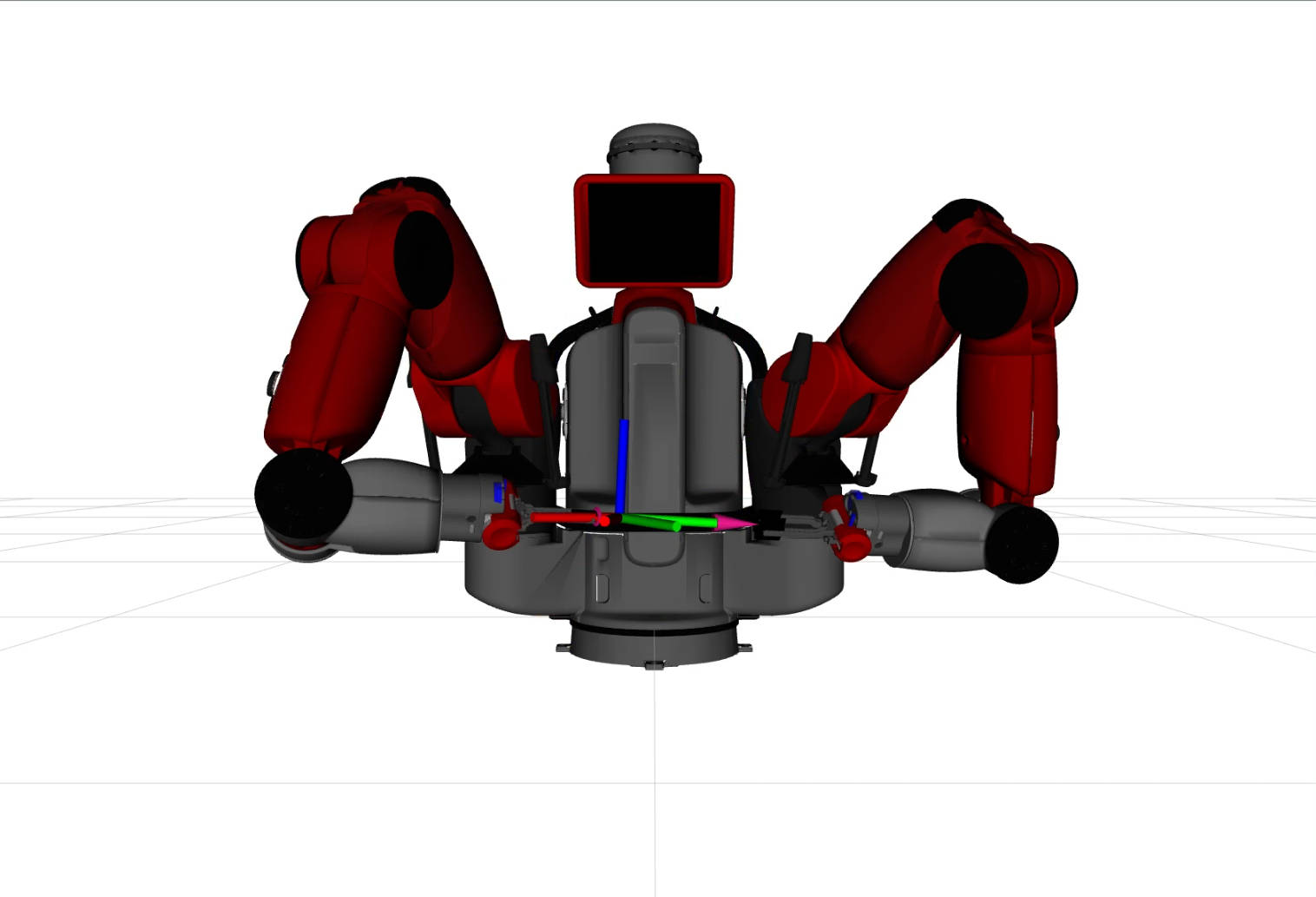}
		\caption{\label{rot_front}}
	\end{subfigure}
	\quad
	\begin{subfigure}[b]{\robwidth}
		\includegraphics[width = \textwidth, trim={200, 200, 100, 100}, clip]{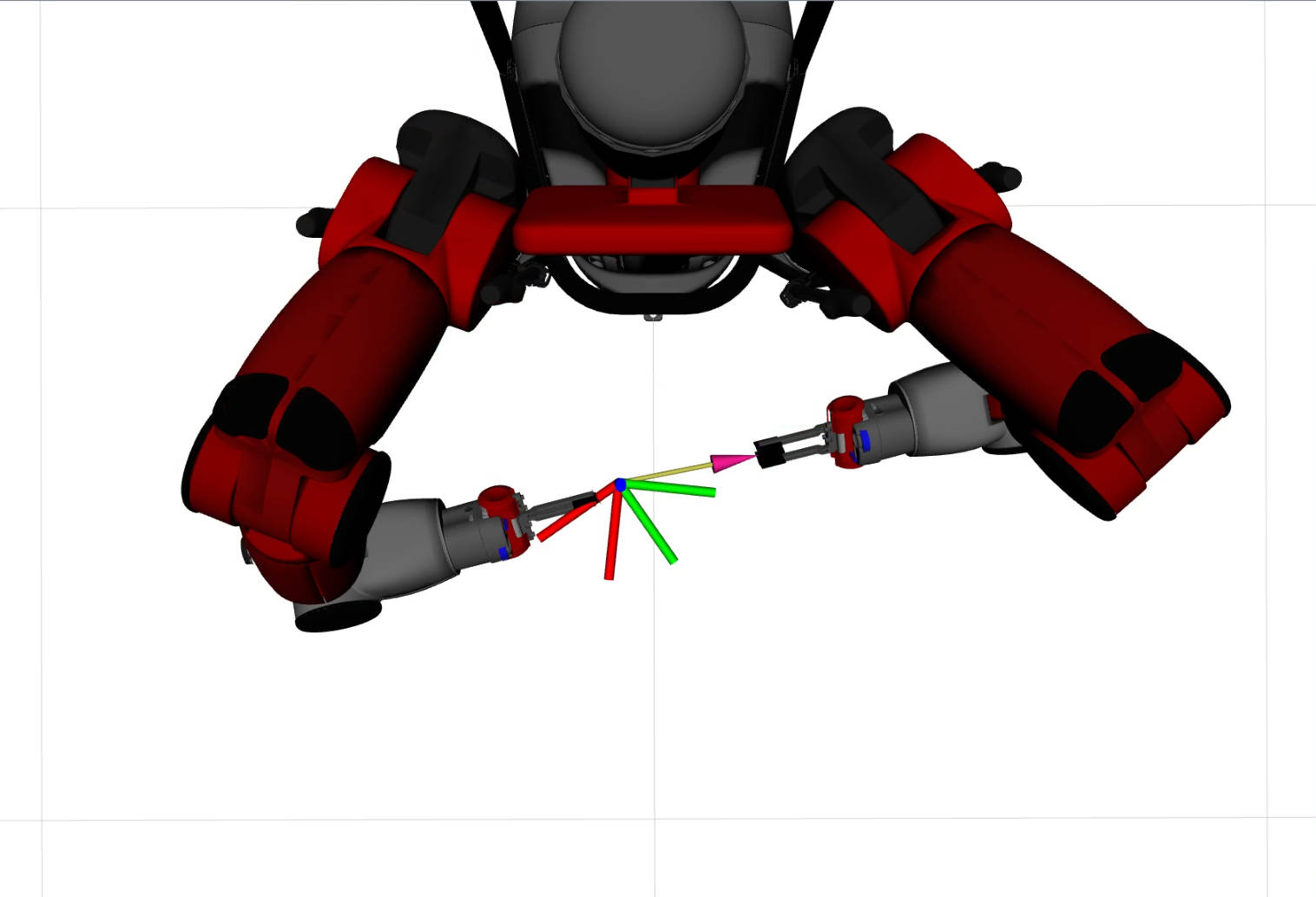}
		\caption{\label{rot_top}}
	\end{subfigure}
	\caption{Initial robot configurations for the case studies. Fig. \ref{linear_front} and \ref{linear_top}: linear displacement between frames seen from a front and a top view. Fig. \ref{rot_front} and \ref{rot_top}: rotational displacement seen from a front and a top view. \label{initia_config}}
\end{figure}
A common choice for $\boldsymbol{\zeta}$  is to specify a motion for one of the system's end-effectors \cite{ Ajoudani2014,Jamisola2015, Hu2015, Foresi2017}, e.g.,
\begin{equation}
	\boldsymbol{\zeta} = [\mathbf{J}_1\; \mathbf{0}]^\dagger \mathbf{v}_1.
	\label{master_slave_sec}
\end{equation}
As seen in Example \ref{example_mapping}, this secondary task will be mapped into the system's absolute motion space and induce asymmetries in the execution of $\mathbf{v}_r$. 
We propose instead to use the asymmetric relative motion space defined in section \ref{asymmetric_def} to extend the relative Jacobian methods and allow for setting the degree of cooperation through a parameter.
We use \eqref{asymmetric_linking} to define an asymmetric Jacobian, 
\begin{equation}
	\mathbf{J}_r(\alpha) = \boldsymbol{\mathcal{L}}_r(\alpha)\mathbf{W}\mathbf{J}.
	\label{asym_jacobian}
\end{equation}
 We can now solve the differential IK by defining
\begin{equation}
	\boldsymbol{\zeta} = \mathbf{J}_{r}(\alpha)^\dagger\mathbf{v}_r.
	\label{asym_relative_jac}
\end{equation}
Imposing \eqref{asym_relative_jac} as a secondary task will filter out undesired asymmetric absolute motion, which can occur as part of the minimum norm solution represented by the pseudoinverse, since it belongs to an orthogonal motion space,  eq. \eqref{asymmetric_orthogonality}.
The desired task space relative motion will be preserved, however, as shown in Corollary \ref{asym_corollary}, and as we will illustrate in Sec. \ref{case_studies}.
Note that asymmetric absolute motion necessarily contains relative motion components, which is in general undesirable: for many tasks, non-prescribed relative motion can lead to internal stresses on the robot manipulator.

\begin{figure}[t]
	\centering
	\begin{subfigure}[t]{0.475\textwidth}
		\centering
	 	\includegraphics[width = \textwidth]{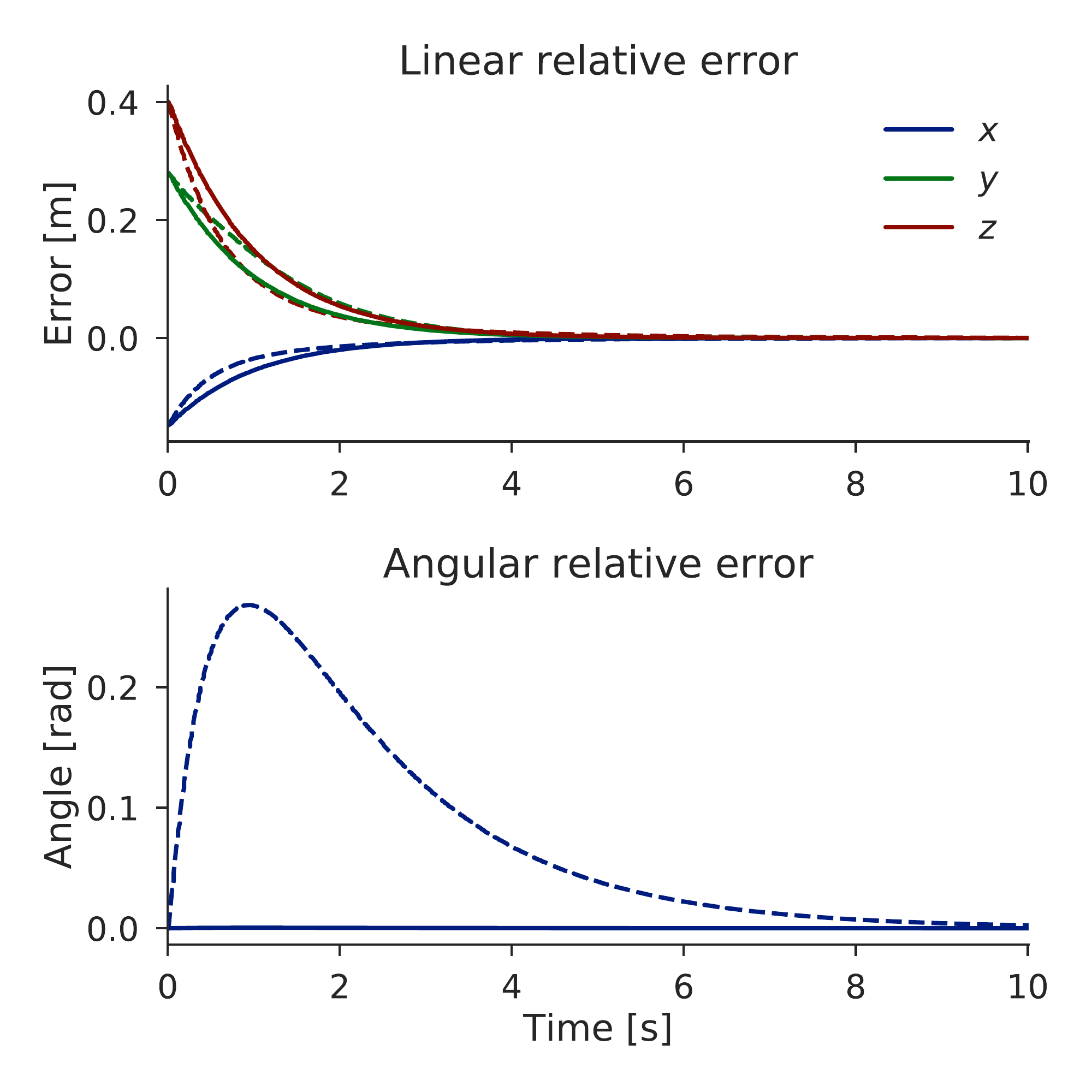}
		\caption{\textbf{Translational} relative motion task. \label{asymmetric_trans_task}}
	\end{subfigure}
	\quad
	\begin{subfigure}[t]{0.475\textwidth}
		\centering
 		\includegraphics[width = \textwidth]{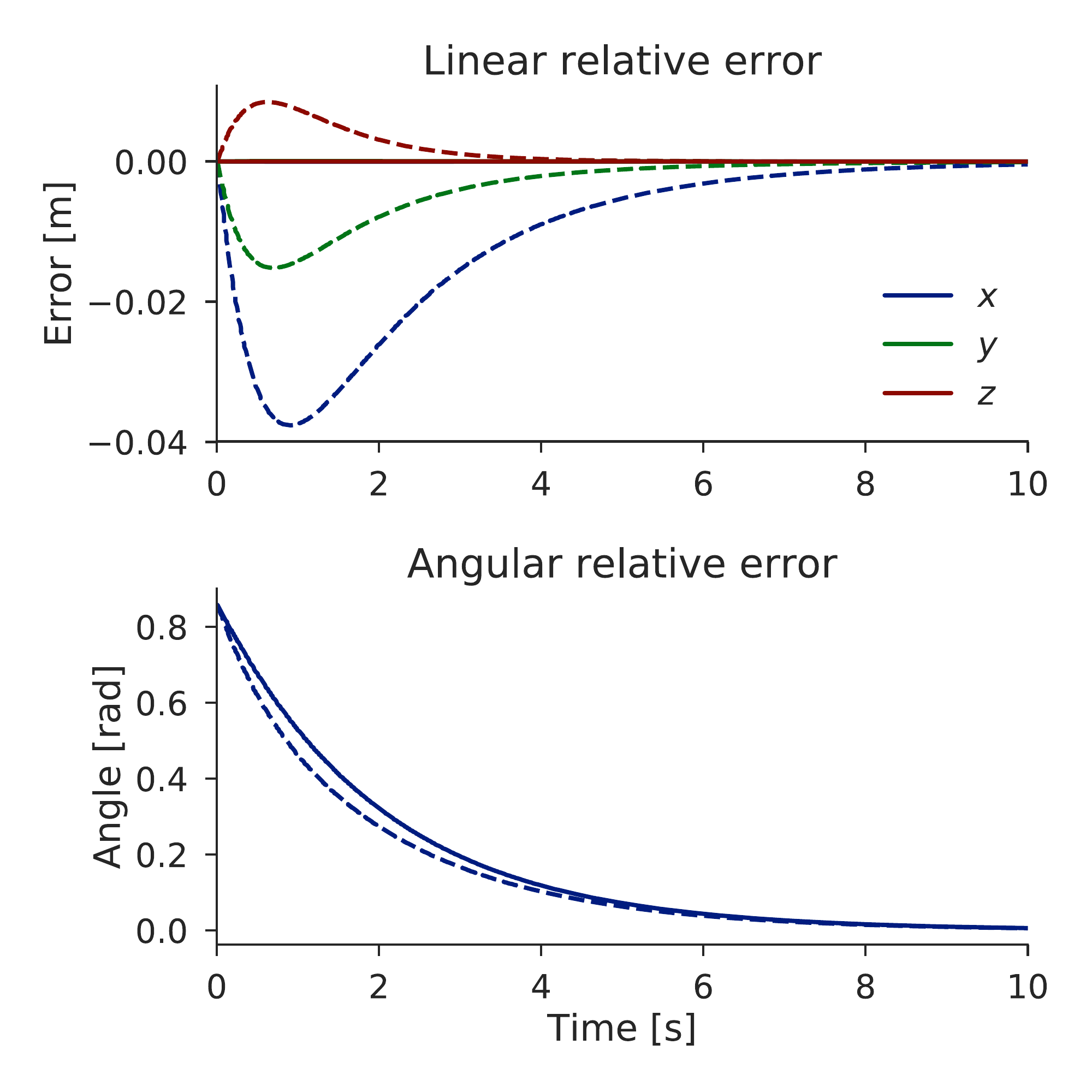}
		\caption{\textbf{Rotational} relative motion task. \label{asymmetric_rot_task}}
	\end{subfigure}
	\caption{Evolution of the relative errors in the two types of motion task, when $\alpha = 0.8$. The dashed lines represent the solution when using $\boldsymbol{\zeta}$ from \eqref{asym_relative_jac} as a primary task. The solid lines use \eqref{asym_relative_jac} in a task priority manner \eqref{ik}.}
\end{figure}
\subsection{Relative task}
The choice of $\mathbf{v}_r$ in \eqref{ik} defines the relative motion task. 
This can be set as the output of a task-specific controller \cite{Almeida2018}, a feedforward command from a teleoperator, or as an error signal from a pose regulation task \cite{Caccavale2000, Park2016}.
In our examples, we will assume without loss of generality that the task is to align the coordinate frames $\{h_{o_i}\}$. 
The alignment error is given by $\tilde{\mathbf{p}} = \mathbf{p}_{o_2} - \mathbf{p}_{o_1}$ for the displacement between frames and $\tilde{\mathbf{R}} = \mathbf{R}_{o_1}^\top \mathbf{R}_{o_2}$ for the orientation. 
If we denote the error quaternion as $\tilde{\mathcal{Q}} = (\tilde{\boldsymbol{\xi}},\,w)$, where $\tilde{\boldsymbol{\xi}}$ is the vector and $w$ the scalar part, we set $\mathbf{v}_r$ as the feedback control law \cite{Caccavale2000},
\begin{equation}
	\mathbf{v}_r = -\mathbf{K}_p \begin{bmatrix}
		\tilde{\mathbf{p}}\\
		\mathbf{R}_{o_1}\tilde{\boldsymbol{\xi}}
	\end{bmatrix},
	\label{relative_task}
\end{equation}
where $\mathbf{K}_p \in \mathbb{R}^{6 \times 6}$ is positive definite.

\section{Case Studies \label{case_studies}}
We illustrate different properties of the novel relative Jacobian \eqref{asym_jacobian} with two case studies, in which the primary task is a relative motion task given by \eqref{relative_task} with $\mathbf{K}_p = \mathbf{I}_6$.
The case studies are based on a simulated Rethink Robotics' Baxter dual-armed robot.
Two relative motion tasks are considered, one where the alignment error is purely translational, Fig. \ref{linear_front}-\ref{linear_top}, the other where the error is purely rotational, Fig. \ref{rot_front}-\ref{rot_top}.
The initial poses of the object frames, expressed in the reference frame (Baxter's torso frame), are given by $\mathbf{p}_{o_1} = [0.36, 0.15, 0.36]^\top$ and $\mathbf{p}_{o_2} = [0.45, 0.0, 0.21]^\top$, with $\mathbf{R}_{o_1} = \mathbf{R}_{o_2} = \mathbf{I}_3$.
The subscript index $i = 1$  corresponds to the left and $i = 2$ to the right manipulator.

\subsection{Nullspace projection of the extended relative Jacobian solution}
\begin{figure}[t]
	\centering
	\begin{subfigure}[b]{0.3\textwidth}
		\centering
		\includegraphics[width = \textwidth, trim={300, 80, 200, 100}, clip]{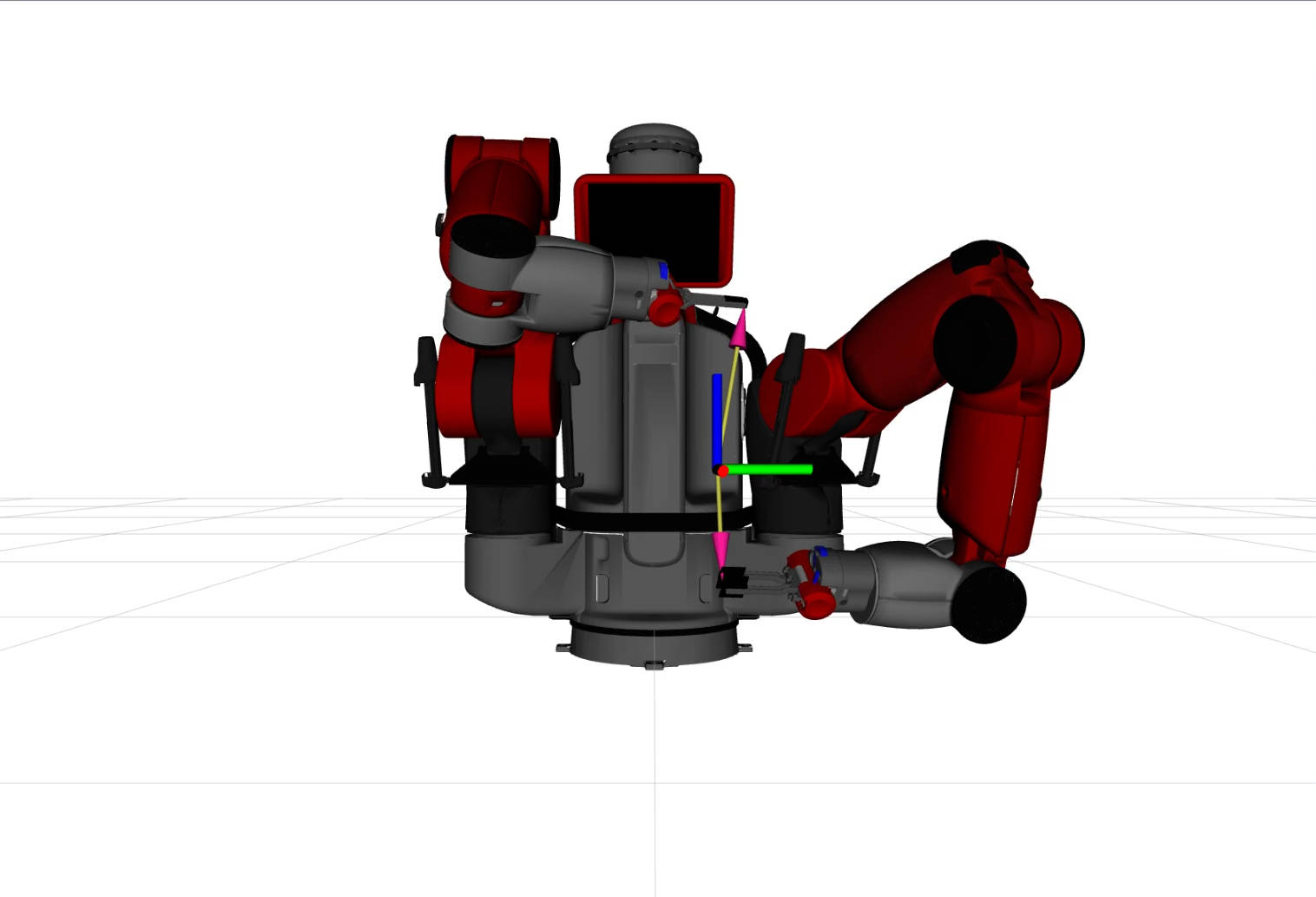}
		\caption{ECTS \label{translational_joints_e08}}
	\end{subfigure}
	\quad
	\begin{subfigure}[b]{0.3\textwidth}
		\centering
		\includegraphics[width = \textwidth, trim={350, 150, 200, 100}, clip]{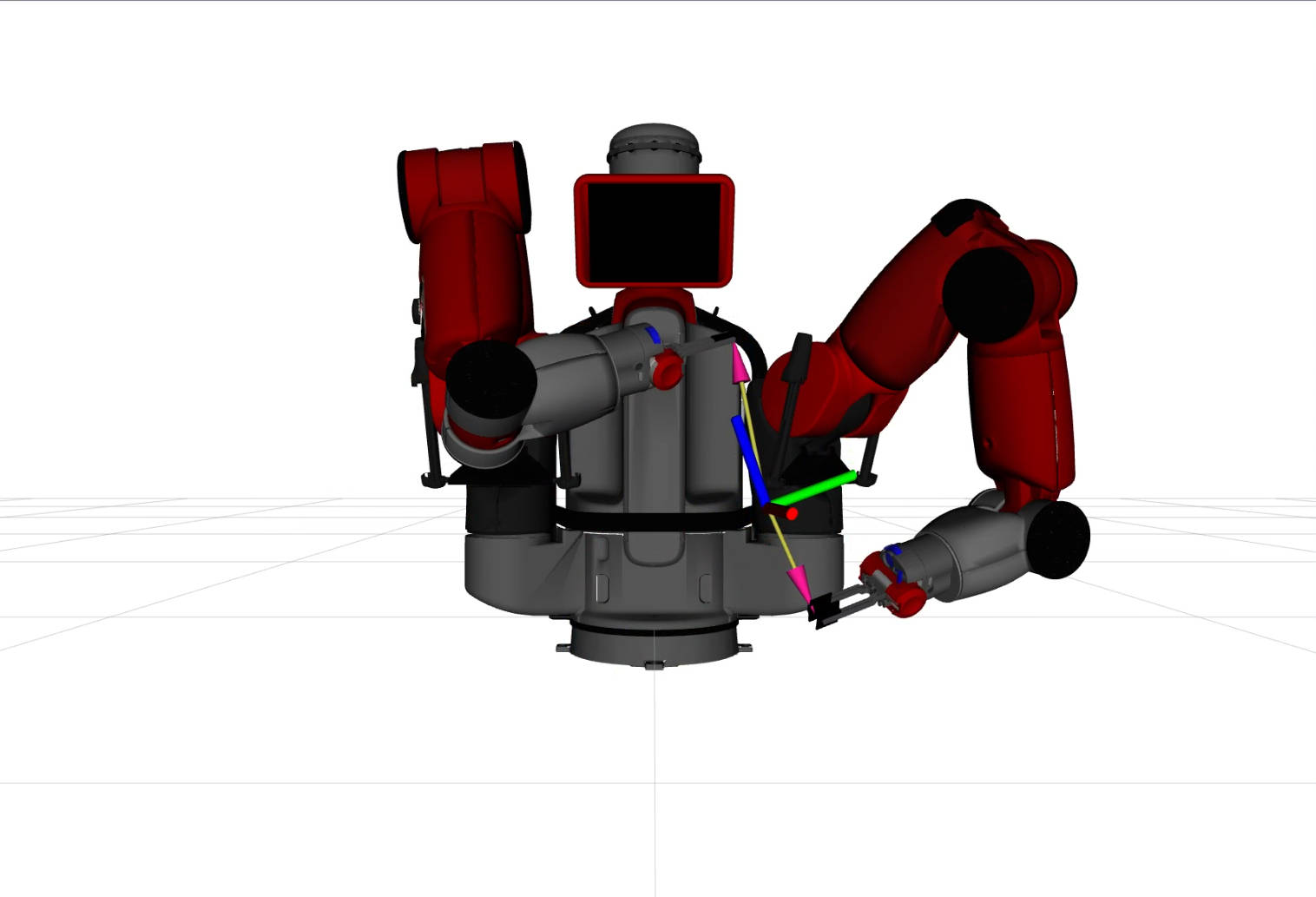}
		\caption{Proposed \label{translational_joints_r08}}
	\end{subfigure}
	\caption{Results for the execution of a pure \textbf{translational} task.\label{translational_joints}}
\end{figure}
\begin{figure}
	\centering
	\begin{subfigure}[b]{0.3\textwidth}
	\centering
		\includegraphics[width = \textwidth, trim={250, 80, 100, 100}, clip]{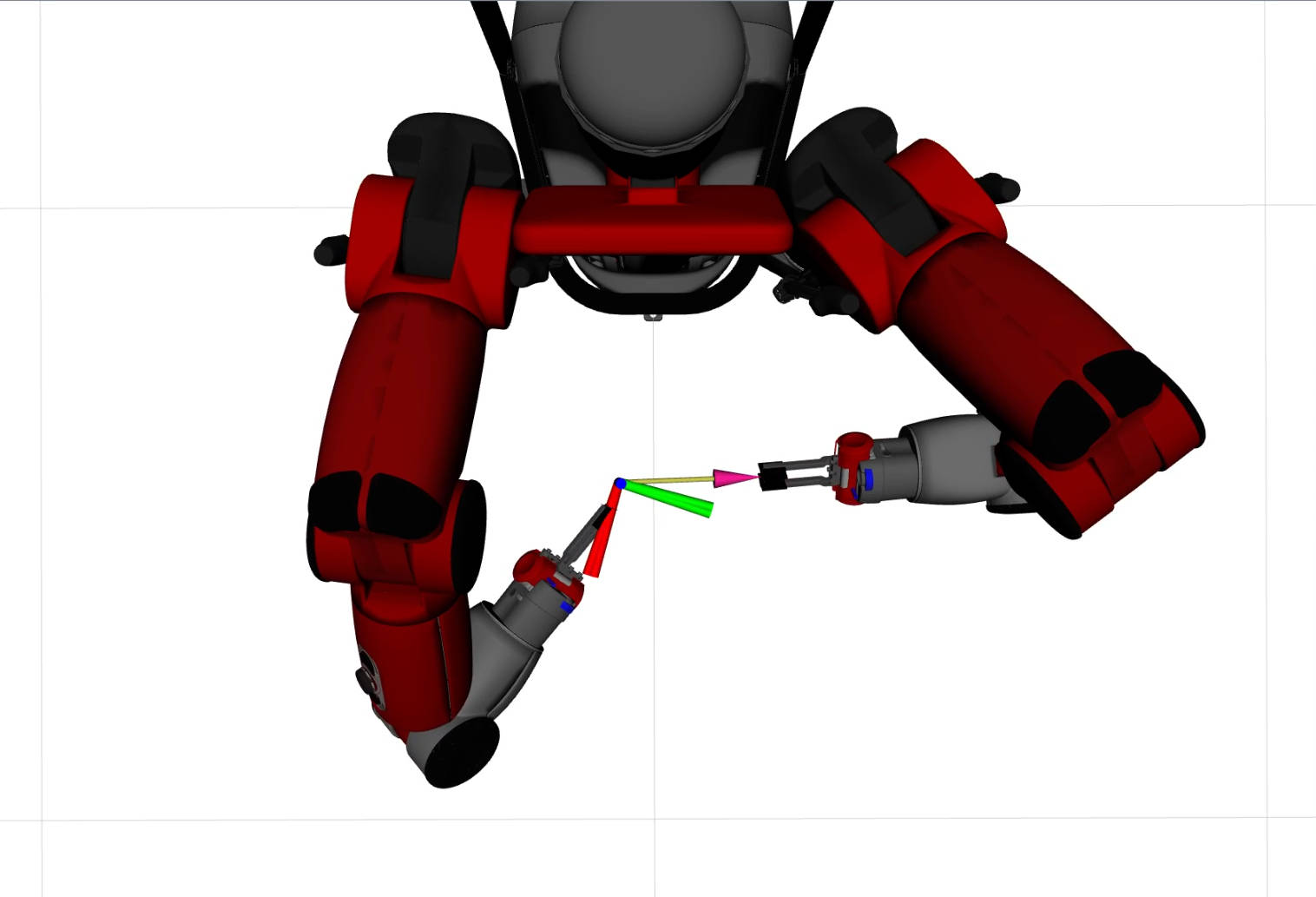}
		\caption{ECTS}
	\end{subfigure}
	\quad
	\begin{subfigure}[b]{0.3\textwidth}
		\centering
			\includegraphics[width = \textwidth, trim={250, 80, 100, 100}, clip]{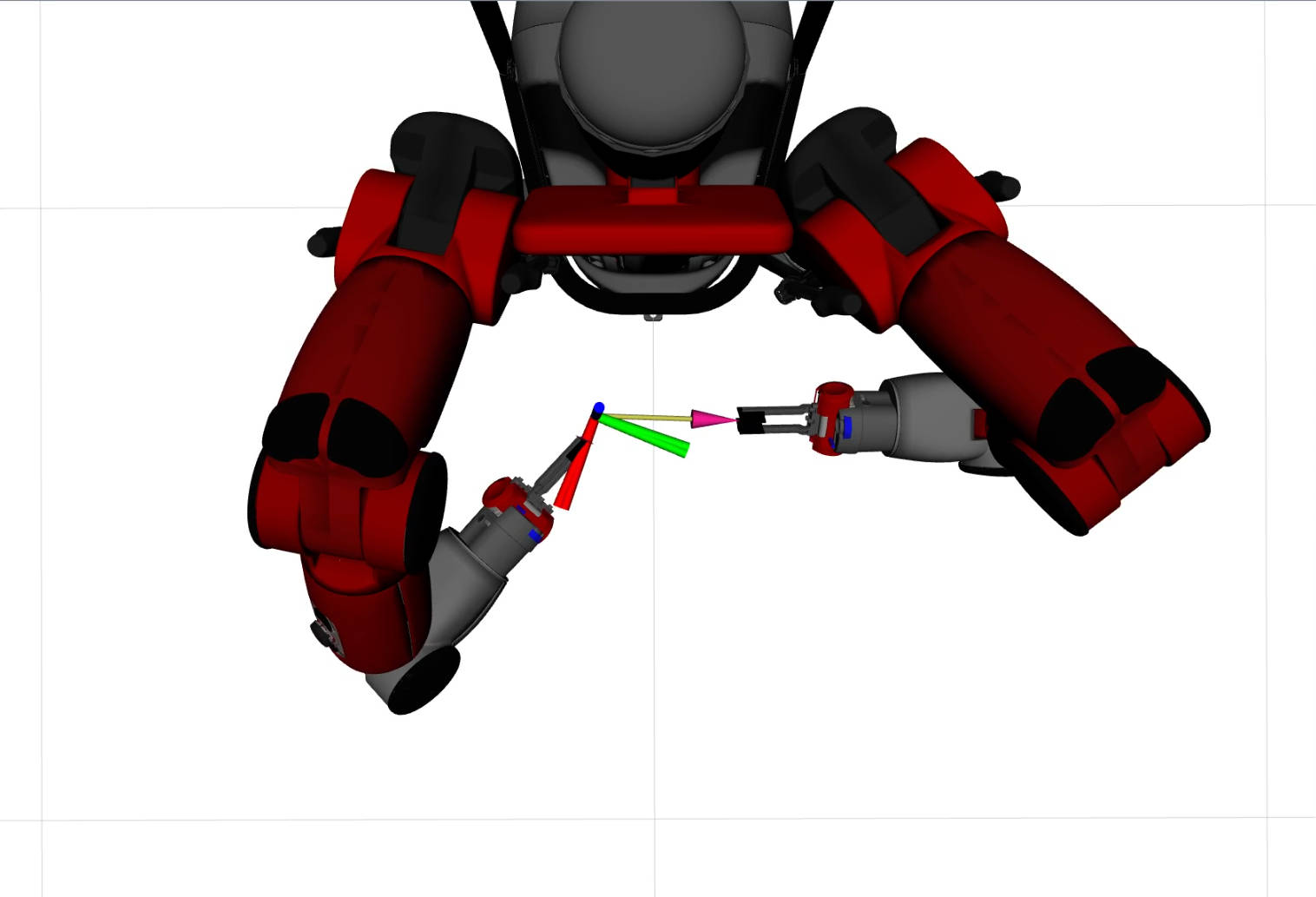}
			\caption{Proposed \label{rotational_joints_r08}}
	\end{subfigure}
	\caption{Results for the execution of a pure \textbf{rotational} task.	 \label{rotational_joints}}
\end{figure}

\noindent In this case study, we show how the joint space solution \eqref{asym_relative_jac} introduces undesirable relative motion components, as a consequence of the asymmetric orthogonality from eq. \eqref{asymmetric_orthogonality}. 
This property implies that assymetric absolute motion lies in the nullspace of \eqref{asym_jacobian}, which contains relative motion components, eq. \eqref{relative_from_abs}.
As such, the nullspace projection \eqref{ik} is required to use our proposed extended relative Jacobian.
In the numerical simulations, we set $\alpha = 0.8$ to denote a blended mode where the right manipulator ($i = 2$) executes most of the relative task.
Fig. \ref{asymmetric_trans_task} depicts the evolution of the relative error for a purely translational task. 
Conversely, Fig. \ref{asymmetric_rot_task} shows the error progression in a purely rotational task.
The solution \eqref{asym_relative_jac} induces undesired relative motion in both tasks. 
This is clearly visible in the rotational error for the translational task and in the translational error for the rotational task. 
The nullspace projection of \eqref{asym_relative_jac} in \eqref{ik} filters out these undesired components.

\subsection{Asymmetric relative motion}
\begin{figure}[t]
	\centering
	\begin{subfigure}[b]{0.45\textwidth}
		\centering
		\includegraphics[width = \textwidth]{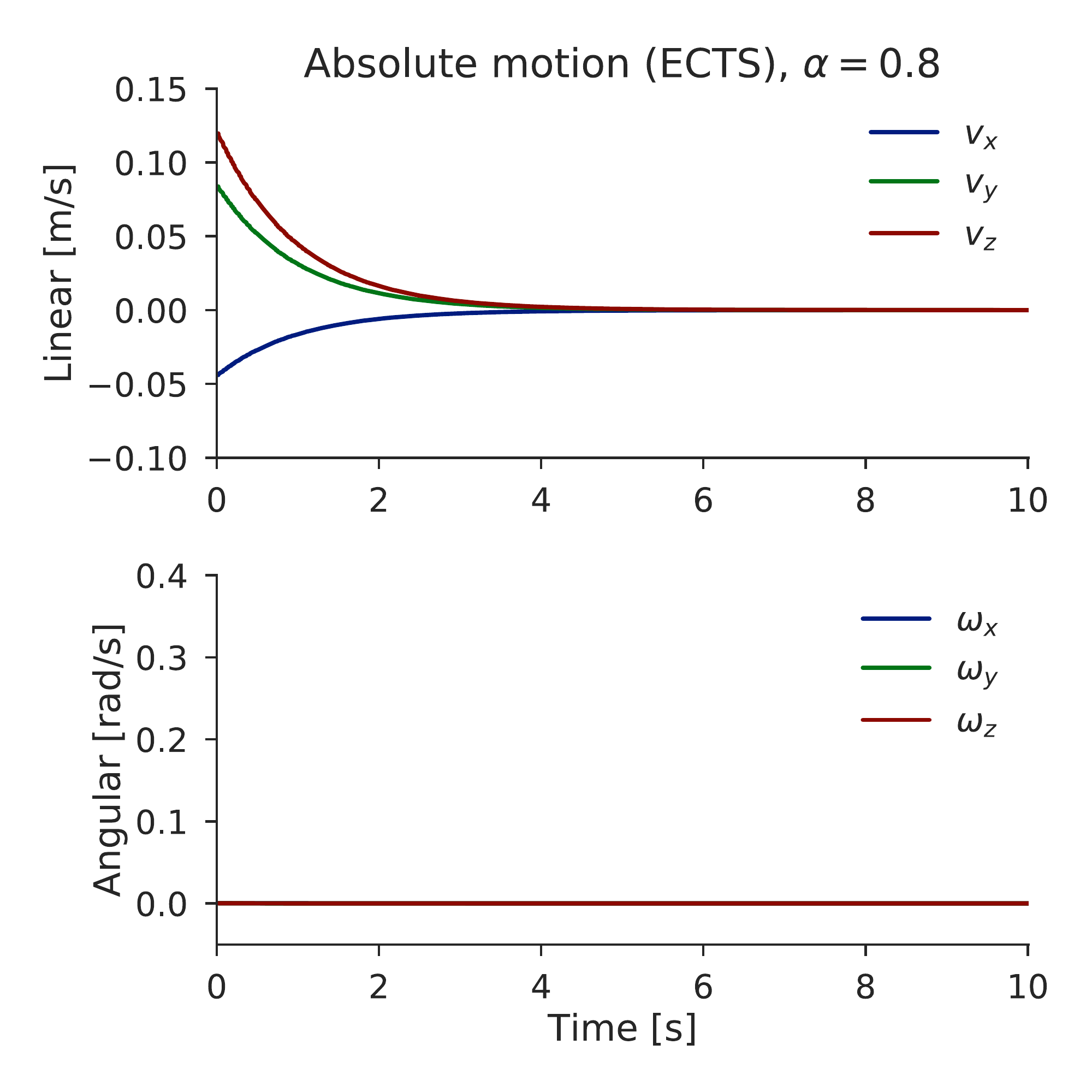}
		\caption{ECTS\label{abs_trans_ects}}
	\end{subfigure}
	\quad
	\begin{subfigure}[b]{0.45\textwidth}
		\centering
		\includegraphics[width = \textwidth]{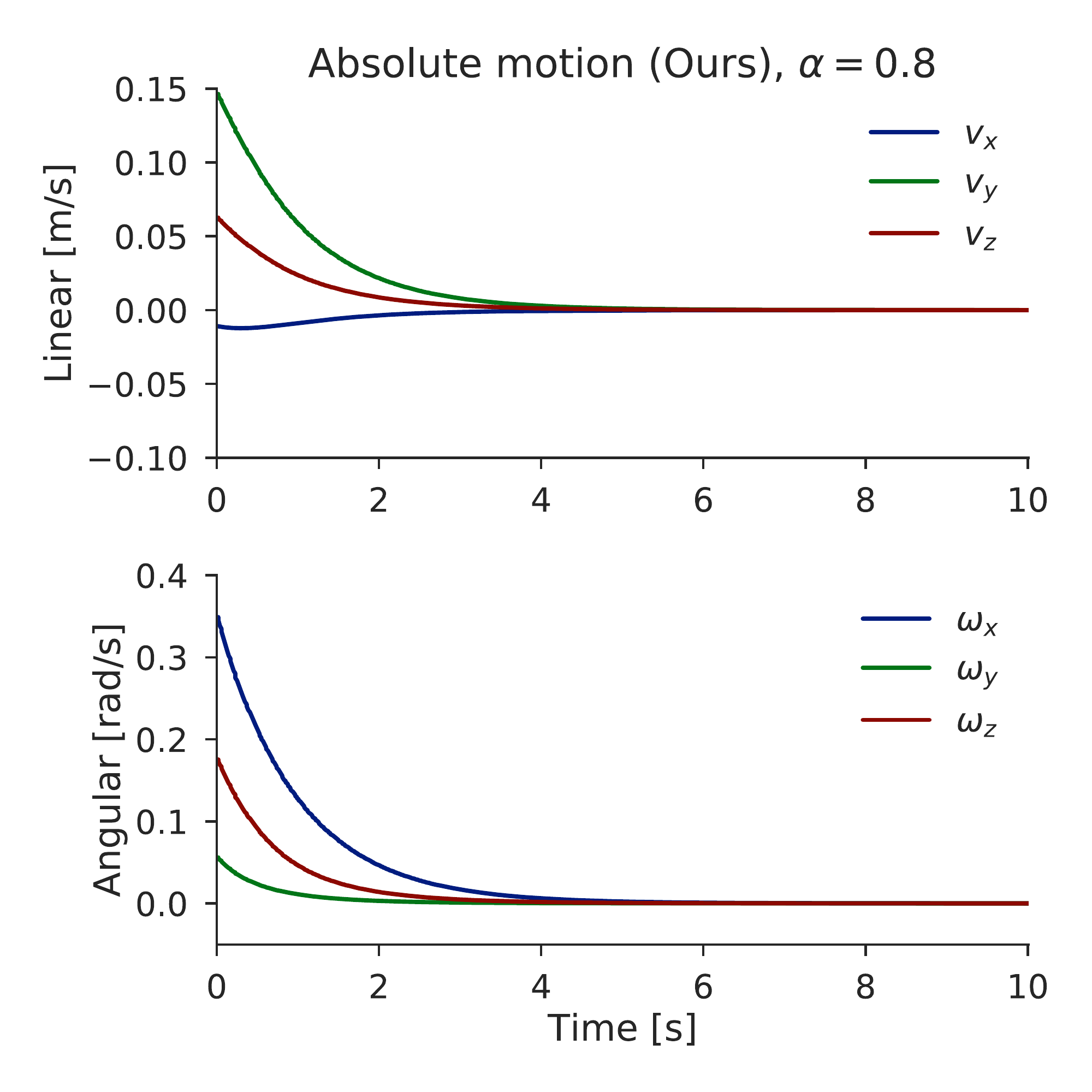}
		\caption{Proposed\label{abs_trans_ours}}
	\end{subfigure}
	\caption{\textbf{Translational} relative task. \label{abs_trans}}
\end{figure}

The nullspace projection of the asymmetric solution \eqref{asym_relative_jac} removes some motion components which are redundant to the asymmetric relative task. 
In this case study, we show that the symmetric absolute motion still remains as an exploitable functional redundancy, and that asymmetric relative motion is achieved, by comparing our method against the ECTS solution with $\mathbf{v}_a = \mathbf{0}$, for different values of the coefficient $\alpha$. 
Note that when $\alpha = 0.5$, we obtain the default CTS and relative Jacobian solutions, respectively.

\begin{figure}[t]
	\centering
	\begin{subfigure}[b]{0.45\textwidth}
		\centering
		\includegraphics[width = \textwidth]{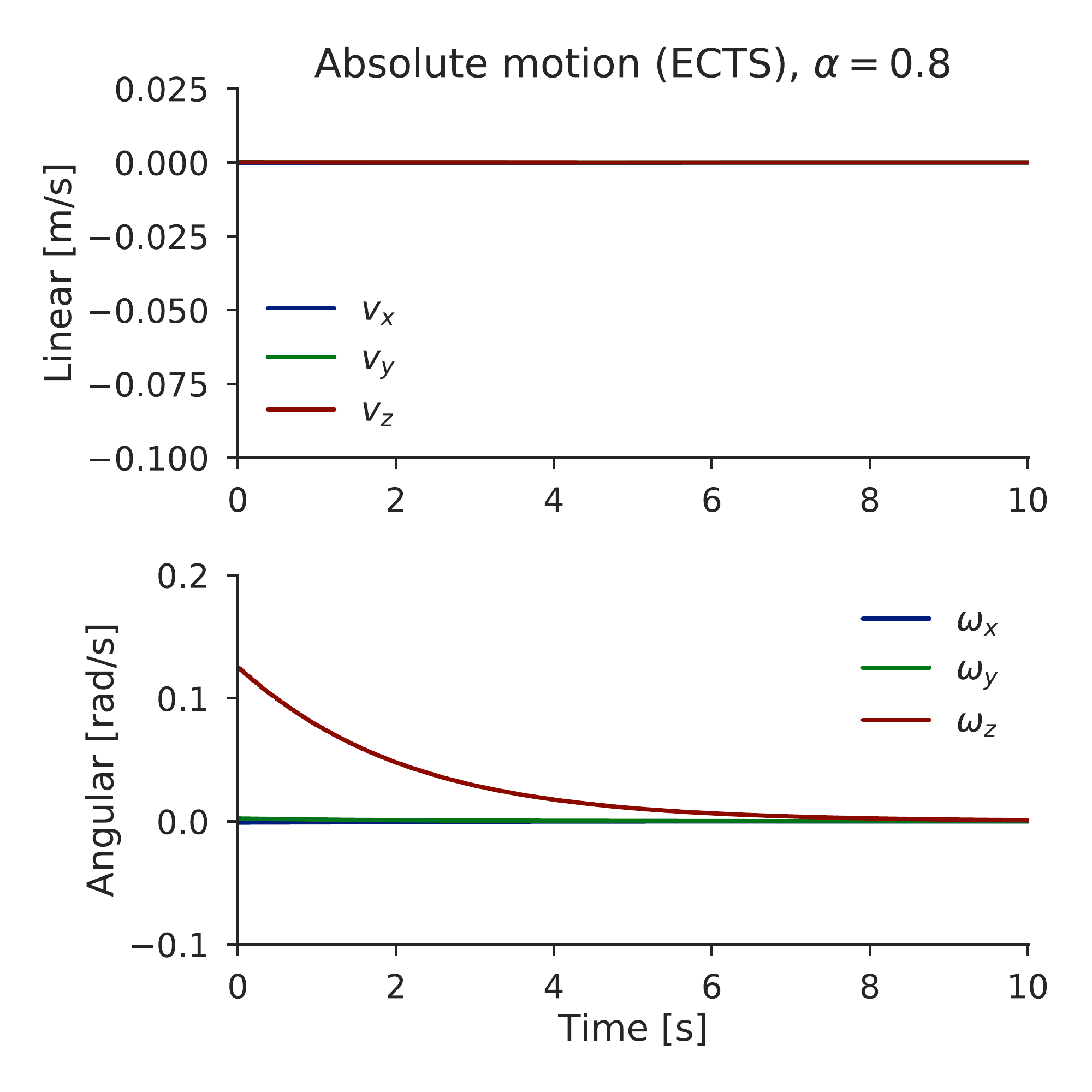}
		\caption{ECTS\label{abs_rot_ects}}
	\end{subfigure}
	\quad
	\begin{subfigure}[b]{0.45\textwidth}
		\centering
		\includegraphics[width = \textwidth]{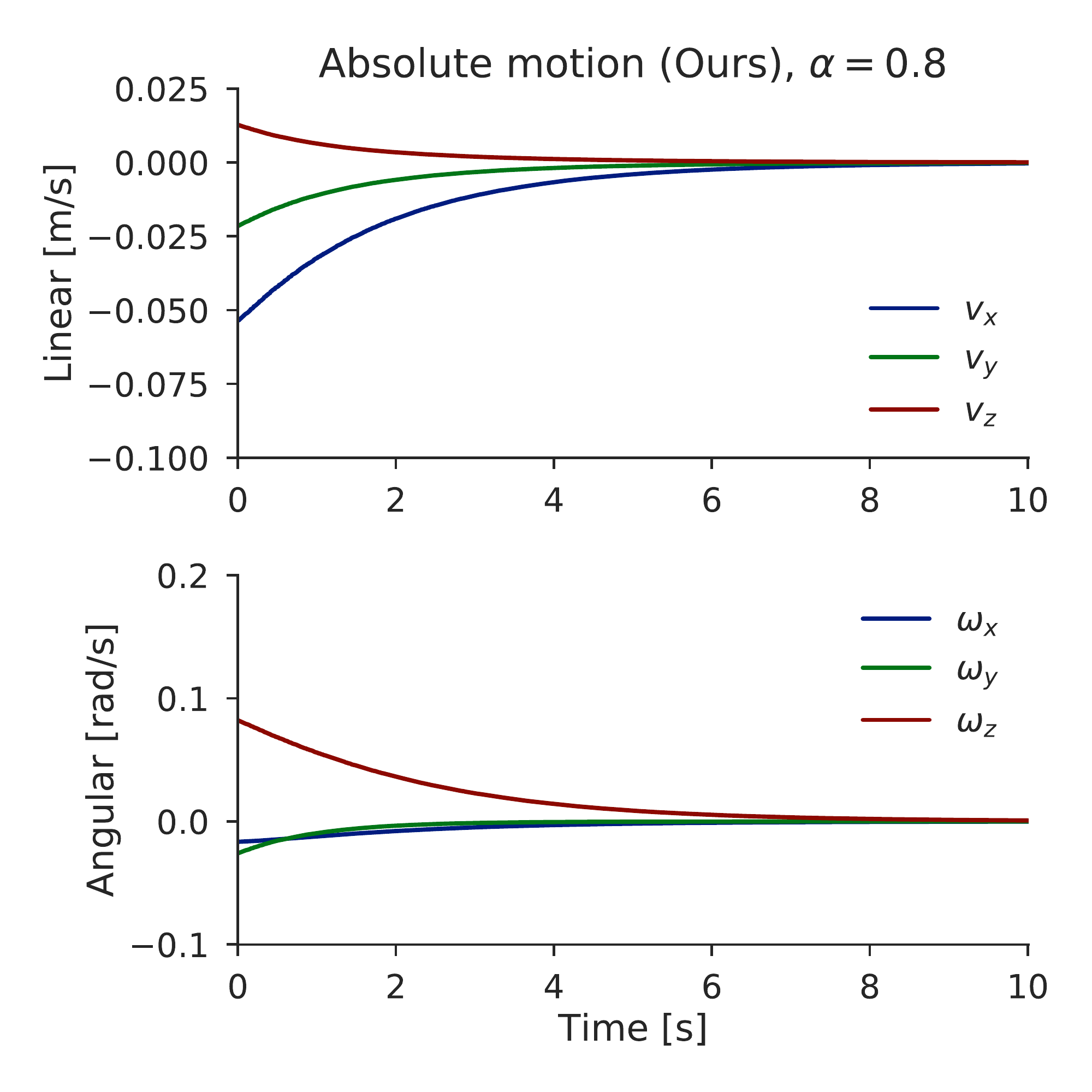}
		\caption{Proposed\label{abs_rot_ours}}
	\end{subfigure}
	\caption{\textbf{Rotational} relative task.\label{abs_rot}}
\end{figure}

We depict the joint solutions when $\alpha = 0.8$ in Fig. \ref{translational_joints}, for the translational relative motion task and in Fig. \ref{rotational_joints} for the rotational task, where it can be observed that the asymmetric task execution results in distinct final configurations for the system.
The norm of the joint space trajectory $\int ||\dot{\mathbf{q}}|| \text{d}t$ was $1.48$ for the ECTS solution and $0.84$ for ours, showing that the availability of the larger redundant space allows for a smaller norm of joint velocities to be computed by the differential IK method.
In addition, we show the induced symmetrical absolute motion in both cases in Fig. \ref{abs_trans} and \ref{abs_rot}.
We provide the source code required to test other values of $\alpha$\footnote{\url{https://github.com/diogoalmeida/asymmetric_manipulation}}.

The smaller joint trajectory norm of our method is a result of the exploitation of the absolute motion as a functional redundancy.
In the translational motion task our method changes the absolute orientation of the system, Fig. \ref{abs_trans_ours}, and conversely a linear component is introduced in the rotational motion task, Fig. \ref{abs_rot_ours}.
In comparison, ECTS will induce symmetrical absolute motion on the system when $\alpha \neq 0.5$, Fig. \ref{abs_trans_ects} and \ref{abs_rot_ects}, however, this is only along the dimensions commanded by $\mathbf{v}_r$, as  noted in eq. \eqref{induced_sym_abs}.

\section{Conclusions}
We study how existing methods to model the differential kinematics of dual-armed robotic systems distribute the relative motion between end-effectors, in the context of cooperative manipulation tasks.
The asymmetric execution of a relative motion task can be achieved by redefining the absolute motion space, as in the ECTS formulation, or through the addition of a secondary task in a relative Jacobian setting.
The ECTS approach requires absolute motion to be part of the system's primary task, and relies on an unintuitive definition of an asymmetric absolute motion space, which is useful solely for the purpose of the asymmetric execution of the relative task.
In a relative Jacobian setting, adding a secondary task induces time-varying asymmetries which are a function of the secondary task's definition, as we illustrated in Example \ref{example_mapping}.

In contrast, we show how a deliberate degree of asymmetry can be imposed on the execution of a relative motion task within a relative Jacobian formulation.
This approach allows for, e.g., a master-slave execution of the relative motion task, as opposed to the time-varying asymmetric execution of other relative Jacobian methods.
We illustrate the properties of this approach with two case studies. 
In the first study, we show that the asymmetric absolute motion which lies in the nullspace of our proposed Jacobian includes undesirable relative motion components, and we illustrate how our proposed differential IK method filters out these components.
Then, in our second case study, we show how the absolute motion space is retained as a functional redundancy even after applying our differential IK scheme, through a comparison with the ECTS approach: for the same relative motion task, our method obtains smaller joint velocity norms, thanks to the induced absolute motion.

\bibliographystyle{spmpsci} 
\bibliography{biblio}
\end{document}